\documentclass[conference]{IEEEtran}
\IEEEoverridecommandlockouts

\usepackage{amsmath,amssymb,amsfonts}
\usepackage{algorithmic}
\usepackage{graphicx}
\usepackage{textcomp}
\usepackage{xcolor}
\usepackage{multirow, makecell}
\usepackage{subcaption}
\usepackage{footnote}
\makesavenoteenv{tabular}
\makesavenoteenv{table}
\usepackage{cite}

\def\BibTeX{{\rm B\kern-.05em{\sc i\kern-.025em b}\kern-.08em
    T\kern-.1667em\lower.7ex\hbox{E}\kern-.125emX}}
\begin{document}

\newcommand{\specialcell}[2][c]{%
  \begin{tabular}[#1]{@{}c@{}}#2\end{tabular}}

\title{\textsc{PermDNN}: Efficient Compressed DNN Architecture with Permuted Diagonal Matrices\\
\thanks{*Chunhua Deng and Siyu Liao contribute equally to the paper.}
\thanks{+Chunhua Deng, Siyu Liao, Yi Xie and Bo Yuan are now with Rutgers University.}
}


 \author{
 \IEEEauthorblockN{Chunhua Deng$^{*+}$} \IEEEauthorblockA{\textit{City University of New York} \\chunhua.deng@rutgers.edu}\\
 \IEEEauthorblockN{Keshab K. Parhi}
 \IEEEauthorblockA{\textit{University of Minnesota, Twin Cities}\\parhi@umn.edu}
 \and
 \IEEEauthorblockN{Siyu Liao$^{*+}$}
 \IEEEauthorblockA{\textit{City University of New York} \\sl1583@scarletmail.rutgers.edu}\\
 \IEEEauthorblockN{Xuehai Qian}
 \IEEEauthorblockA{\textit{University of Southern California}\\
 xuehai.qian@usc.edu}
 \and
 \IEEEauthorblockN{Yi Xie$^{+}$}
 \IEEEauthorblockA{\textit{City University of New York} \\yx238@scarletmail.rutgers.edu}\\
 \IEEEauthorblockN{Bo Yuan$^{+}$}
 \IEEEauthorblockA{\textit{City University of New York}\\
 bo.yuan@soe.rutgers.edu}
 }

\maketitle

\begin{abstract}
Deep neural network (DNN) has emerged as the most important and popular artificial intelligent (AI) technique. 
The growth of model size poses a key energy
efficiency challenge for the underlying
computing platform.
Thus, model compression becomes a crucial problem.
However, the current approaches
are limited by various drawbacks.
Specifically, network sparsification approach
suffers from irregularity, heuristic nature and large indexing overhead.
On the other hand, the recent
 structured matrix-based approach (i.e., \textsc{CirCNN})
 is limited by 
the relatively complex arithmetic computation 
(i.e., FFT), less flexible compression ratio, 
and its inability to fully utilize input sparsity.

To address these drawbacks, this paper proposes \textsc{PermDNN}, a novel approach to generate and execute hardware-friendly structured sparse DNN models using permuted diagonal matrices. Compared with unstructured sparsification approach, \textsc{PermDNN} eliminates the drawbacks of indexing overhead, non-heuristic compression effects and time-consuming retraining. Compared with circulant structure-imposing approach, \textsc{PermDNN} enjoys the benefits of higher reduction in computational complexity, flexible compression ratio, simple arithmetic computation and full utilization of input sparsity.
We propose \textsc{PermDNN} architecture, a multi-processing element (PE) fully-connected (FC) layer-targeted computing engine. The entire architecture is highly scalable and flexible, and hence it can support the needs of different applications with different model configurations. 
We implement a 32-PE design using CMOS 28nm technology. Compared with EIE, \textsc{PermDNN} achieves $3.3\times\sim4.8\times$ higher throughout, $5.9\times\sim8.5\times$ better area efficiency and $2.8\times\sim4.0\times$ better energy efficiency on different workloads. Compared with \textsc{CirCNN}, \textsc{PermDNN} achieves $11.51\times$ higher throughput and $3.89\times$ better energy efficiency.
\end{abstract}


\section{Introduction}
\label{sec:intro}

Starting their resurgence from Hinton's seminal paper \cite{hinton2006reducing}, neural networks \cite{Goodfellow-et-al-2016} have emerged as today's most important and powerful artificial intelligence (AI) technique. Thanks to the availability of unprecedentedly abundant training/test data and the significant advances in computers' processing speed, large-scale deep neural networks (DNNs) have been able to deliver record-breaking accuracy results in many tasks that demand intelligence, such as speech recognition \cite{deng2013new}, object recognition \cite{deng2009imagenet}, natural language processing \cite{collobert2008unified} etc.

The extraordinary performance of DNNs with respect to high accuracy is mainly attributed to their very large model sizes \cite{coates2013deep,schmidhuber2015deep,le2013building}. As indicated in a number of theoretical analysis \cite{raghu2016expressive}\cite{liang2016deep} and empirical simulations \cite{he2016deep,zagoruyko2016wide,huang2017densely}, scaling up the model sizes can improve the overall learning and representation capability of the DNN models, leading to higher classification/predication accuracy than the smaller models. Motivated by these encouraging findings, the state-of-the-art DNNs continue to scale up with the purpose of tackling more complicated tasks with higher accuracy.

On the other hand, from the perspective of hardware design, the continuous growth of DNN model size poses a key energy efficiency challenge for the 
underlying computing platforms. As pointed out in \cite{han2016eie}, since the size of on-chip SRAM is usually very limited, placing the large-scale DNN models on the off-chip DRAM, which has more than 100 times higher energy cost than SRAM, is a bitter but inevitable choice. Consequently, high energy cost incurred by the frequent access to DRAM makes the energy-efficient deployment of DNN systems very challenging.

To improve energy efficiency, efficient 
model compression has emerged as a very 
active topic in AI research community. 
Among different types of DNN compression techniques \cite{han2015learning,liu2015sparse,jaderberg2014speeding,wen2016learning}, two approaches show promising results. 
First, \textit{network sparsification} is believed to be the most popular and state-of-the-art strategy because of its good balance between compression ratio and test accuracy. To date, many methods \cite{han2015learning}\cite{liu2015sparse}\cite{wen2016learning} have been proposed to perform efficient sparsification on different DNN models. 
Echoing the importance and popularity of 
this approach, several sparse model-oriented hardware architectures \cite{chen2016eyeriss,parashar2017scnn,albericio2016cnvlutin,yu2017scalpel} have been proposed .

However, 
the current network sparsification methods suffer from the inherent drawbacks of \textit{irregularity}, \textit{heuristic nature} and \textit{indexing overhead}. Consequently, 
despite the encouraging compression ratio,
the {\em unstructured}
sparse DNN models cannot achieve optimal 
performance on the current 
computing platforms, especially DNN hardware
accelerators.
Essentially, the inefficient execution and 
hardware implementation are caused by 
the {\em non-hardware-friendly} sparse models.

To overcome the drawbacks of the irregularity, 
an alternative approach
is to directly represent the network with 
{\em structured matrices} \cite{pan2012structured}.
A notable recent work is \textsc{CirCNN} \cite{ding2017c}, 
which represents weights using block-circulant matrices
and partitions the weight matrix $\textbf{W}\in \mathbb{R}^{m\times n}$ into blocks of 
square circulant sub-matrices (e.g., $\mathbf{W}_{ij}$ of size $k \times k$). 
Due to circulant structure, only $k$ (instead of $k^2$) weights need to be stored for each sub-matrix.
The calculation of $\mathbf{W}_{ij} \mathbf{x}_j$ can be performed as $\text{IFFT}\big(\text{FFT}(\mathbf{w}_{ij})\circ\text{FFT}(\mathbf{x}_j)\big)$, where $\circ$ denotes element-wise multiplications and $\mathbf{w}_{ij}$ is the first row of $\mathbf{W}_{ij}$.
Utilizing FFT-based fast multiplication,  
it simultaneously reduces 
computational cost and storage cost with small accuracy loss.

Due to the hardware-friendly model representation,
\textsc{CirCNN} demonstrated the efficient hardware implementation and better performance.
However, it is still limited by 
the relatively complex arithmetic computation 
(i.e., FFT), less flexible compression ratio, 
and inability to fully utilize input sparsity. 

\begin{figure}
\centering
\begin{subfigure}{0.4\textwidth}
\centering
{\includegraphics[width=0.8\linewidth]{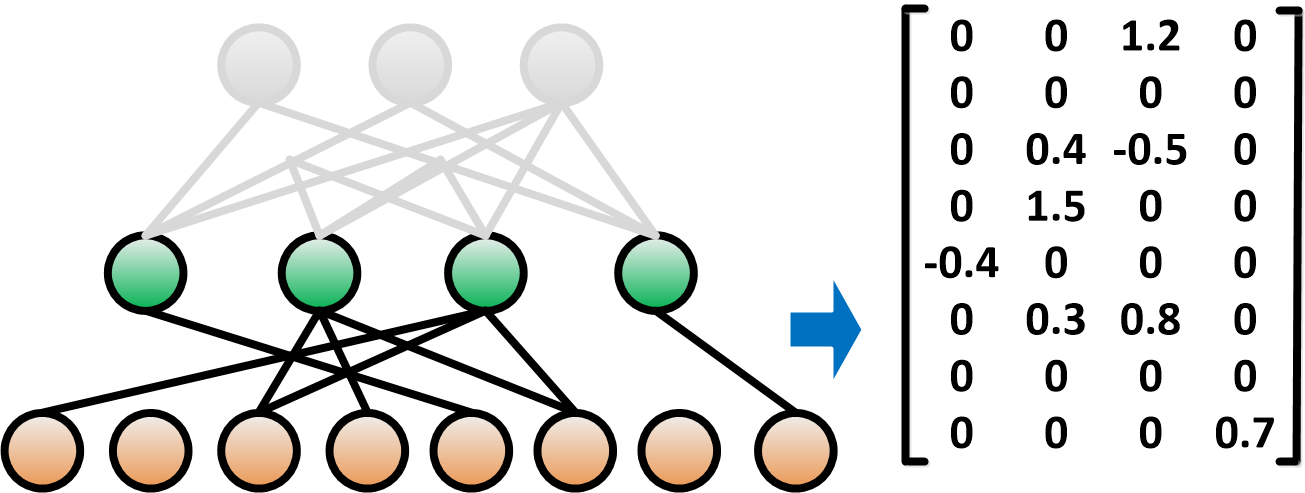}
\caption{Unstructured sparse weight matrix.}
\vspace{0mm}
}
\end{subfigure}
\\
\begin{subfigure}{0.4\textwidth}
{\includegraphics[width=1\linewidth]{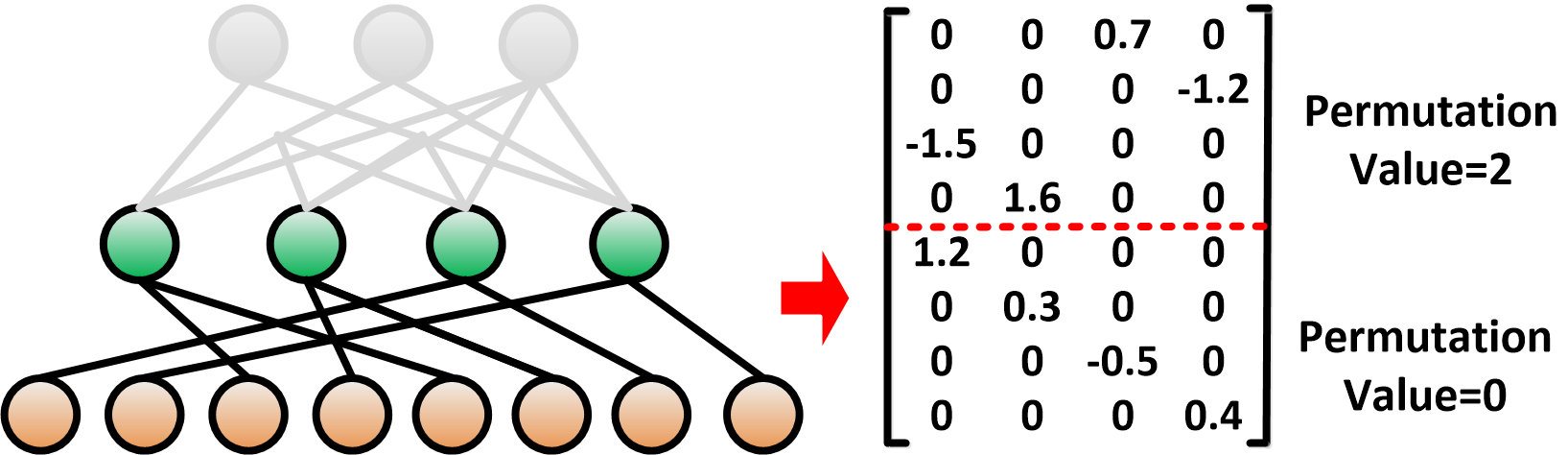}
\caption{Block-permuted diagonal weight matrix.}
\vspace{0mm}
}
\end{subfigure}
\caption{Weight representation by using (a) conventional unstructured sparse matrix. (b) block-permuted diagonal matrix.}
\label{fig:bpd_matrix}
\vspace{-7mm}
\end{figure}

In this paper, we propose \textsc{PermDNN}, a novel approach that generates and executes hardware-friendly structured sparse DNN models using \textit{permuted diagonal matrices}. 
As illustrated in Fig. \ref{fig:bpd_matrix}(b), permuted diagonal matrix is a type of structured sparse matrix that places all the non-zero entries in the diagonal or permuted diagonal. 
When the weight matrices of DNNs can be represented in the format of multiple permuted diagonal matrices (so-called \textit{block-permuted diagonal matrices}), their inherent strong structured sparsity leads to great benefits for practical deployment.
Specifically, it eliminates indexing overhead, brings non-heuristic compression effects, and enables re-training-free model generation.

While both methods impose the structure on the construction of weight matrices,
\textsc{PermDNN} offers three advantages over
\textsc{CirCNN}:
1) \textit{Simpler arithmetic computation}. Unlike \textsc{CirCNN}, which inherently requires complex number operation (complex multiplication and addition) because of the use of FFT, \textsc{PermDNN} is purely based on real number arithmetic, thereby leading to low hardware cost with the same compression ratio; 2) \textit{Significantly better flexibility}. \textsc{PermDNN} hardware can freely support different sizes of permuted diagonal matrices, while \textsc{CirCNN} hardware is limited to only support $2^t$-size circulant matrix because most of FFT hardware is $2^t$-point-based. This means that \textsc{PermDNN} is much more flexible for the choice of compression ratio; 3) \textit{Full utilization of input sparsity}. \textsc{PermDNN} can fully utilize the dynamic sparsity in the input vectors but \textsc{CirCNN} cannot.
The reason is that \textsc{CirCNN} can only process frequency-domain input vectors that lose important time-domain sparsity.
It allows \textsc{PermDNN} to achieve additional improvements in throughput and energy efficiency
over \textsc{CirCNN}.

Based on \textsc{PermDNN}, 
we develop an end-to-end training scheme that can generate a high-accuracy permuted diagonal matrix-based DNN models from scratch. 
We also develop the corresponding low-complexity inference scheme that executes efficiently 
on the trained structured sparse DNN models. Experimental results on different datasets for different application tasks show that, 
enforced with the strong structure, the permuted diagonal matrix-based DNNs achieve high sparsity ratios with no or negligible accuracy loss.

To accelerate inference, we propose
\textsc{PermDNN} architecture, a high-performance permuted diagonal matrix-based inference engine that targets the fully-connected (FC) layers of DNN models. The \textsc{PermDNN} architecture is designed to fully reap the benefits of permuted diagonal matrix-based structure. Unlike EIE \cite{han2016eie}, the state-of-the-art DNN architecture 
targeting FC layer, \textsc{PermDNN} architecture 
does not incur index overhead
and load imbalance due to the 
irregularity of conventional unstructured sparse DNN models, thereby leading to significant improvement in hardware performance.
With an array of processing elements (PEs), \textsc{PermDNN} architecture is an elastic and scalable architecture that can adapt to different
sizes of DNN models, size of component permuted diagonal matrices, and number of PEs.
This allows the architecture to be 
deployed in various application scenarios that have different requirements on power, area and throughput.

To demonstrate the advantages of \textsc{PermDNN} architecture, we implement a 32-PE design using CMOS 28nm technology. Operating on 1.2GHz clock frequency, the \textsc{PermDNN} 
implementation consumes 703.4mW and 8.85mm$^2$. Meanwhile, equipped with 8 multipliers in each PE, the processing power of 32-PE \textsc{PermDNN} achieves 614.4GOPS for a compressed DNN model, which approximately corresponds to 14.74TOPS on an uncompressed network. Compared to EIE, \textsc{PermDNN} achieves $3.3\times\sim4.8\times$ 
higher throughout, $5.9\times\sim8.5\times$ better area efficiency and $2.8\times\sim4.0\times$ better energy efficiency on different workloads.
Compared to \textsc{CirCNN}, \textsc{PermDNN} achieves $11.51\times$ higher throughput and $3.89\times$ better energy efficiency.



\section{MOTIVATION}
\label{sec:motivation}

\subsection{Importance of FC Layer-targeted Architecture}
\label{subsec:importance of fc hw}

Similar to EIE, \textsc{PermDNN} is a customized DNN architecture that targets for FC layers. Specifically, \textsc{PermDNN} accelerates the sparse matrix-vector multiplication ($M\times V$), which is the kernel computation of any FC layers. 
We justify the importance of FC layer 
with the following three arguments.

First, FC layer has a wide spectrum of application scenarios. As summarized in Table \ref{table:FC_applications}, besides being used in convolutional neural network (CNN) for computer vision tasks, FC layer is also the dominant layer in many other types of DNNs (e.g., recurrent neural network (RNN\footnote{In this paper the FC in RNN specifically means the component weight matrices in RNN.}) and multi-layer perceptrons (MLP) for speech recognition and natural language processing tasks. 
    Therefore, optimizing FC layer can universally
    benefit a full spectrum of AI tasks. 
    
Second, real-world industrial needs call for efficient FC layer-targeted DNN architecture. As revealed in Google’s seminal TPU paper \cite{jouppi2017datacenter}, {\em more than 95\%} workload in Google's datacenters are processed by those FC layer-centered DNNs, such as RNNs and MLPs. Therefore, 
    \cite{jouppi2017datacenter} called for more research efforts on FC layers, which is currently far less active than the effort for convolutional (CONV) layers.
    
Finally, optimized FC layer-targeted architecture can improve the overall performance of CONV layer-centered models (e.g., CNNs).
    As analyzed in \cite{parashar2017scnn}, forcing a CONV layer-targeted architecture to execute FC layers will cause noticeable performance degradation.
    We believe that, to realize an efficient architecture for CNNs with both CONV and FC layers, a  specialized design that optimizes for FC layer is a preferable strategy to maximize hardware performance.
    

\begin{table}[h]
  \caption{Types of DNN models used in practical AI fields.}
  \centering
  \begin{tabular}{|c|c|c|}
    \hline
    \textbf{Applications} & \textbf{DNN Models} & \textbf{Component Layer} \\ \hline
    Computer Vision & CNN & \specialcell{CONV layer (Main) \\  FC layer}  \\ \hline
    Speech Recognition & RNN, MLP & FC layer \\ \hline
    Natural Language Processing & RNN, MLP & FC layer \\
    \hline
  \end{tabular}
  \label{table:FC_applications}
  \vspace{0mm}
\end{table}

\subsection{Drawbacks of Unstructured DNN Sparsification}
\label{subsec:drawback of unstructured}

Due to the large memory requirement and 
the inherent redundancy, 
FC layers are usually compressed with 
network sparsification, such as
heuristic pruning \cite{han2015learning} and conducting certain regularization \cite{wen2016learning}.
The downside of the approach is 
that the unstructured sparse models are not friendly to be implemented in DNN hardware accelerators. 

First, the structure of their generated sparse DNNs is usually highly irregular \cite{han2015learning} or only exhibits weak regularity \cite{wen2016learning}. 
It incurs significant space/computation 
overhead due to indexing the irregular sparse weight matrices.
For instance, in EIE, each weight requires 4-bit virtual weight tag to represent its actual value and additional 4 bits to record its relative position in the entire weight matrix. 
Therefore, the overall storage cost for one weight is actually 8 bits instead of 4 bits,
significantly limiting the achievable performance.

Second, the compression ratio of DNN sparsification is typically heuristic. This is because the inherent process of weight/neuron pruning or regularization-caused sparsification  is usually uncontrollable and unpredictable. 
Consequently, in the scenario of compression effect is pre-defined, precisely controlling the compression ratio to satisfy the design specification becomes challenging.

Moreover, the existing DNN sparsification approaches introduce large training overhead. 
Operating on the pre-trained dense model,
weight/neuron pruning and retraining 
are performed iteratively to ensure the original 
accuracy of the dense model can also be achieved by the sparse model.
Clearly, the process is time-consuming,
--- sometimes it may even take longer time to re-train sparse models than the dense models. For instance, as reported in \cite{guo2016dynamic}, in order to get a fair compression rate on a pre-trained AlexNet model, it takes 4800K and 700K iterations by using the pruning methods in \cite{han2015learning} and \cite{guo2016dynamic}, respectively, while training this pre-trained model only takes 450K iterations. 


\vspace{0mm}
\subsection{Drawbacks of Circulant Matrix-based Compression}
\label{subsec:drawback of circnn}

An alternative approach to achieve network
compression is to directly represent the network with {\em structured matrices}.
As an example, \textsc{CirCNN}
utilizes the mathematical property of circulant matrix to compress DNN models that can be 
efficiently implemented in hardware. 
While avoiding the irregularity,
\textsc{CirCNN} suffers from a different set 
of drawbacks.
First, it requires high-cost arithmetic operation. The training and inference algorithms of \textsc{CirCNN} are based on FFT computation, which is inherently involved with complex multiplication and addition. Unfortunately, arithmetic operations on complex numbers incur much higher cost than their counterparts on real numbers. For instance, one complex multiplication requires four real multiplications and two real additions. 

Moreover, it lacks flexibility on compression ratio. The compression ratio of \textsc{CirCNN} is determined by the size of component circulant matrix. However, since this size also determines the length of FFT, in order to facilitate FFT hardware design that mostly uses $2^t$-point FFT\footnote{ non-$2^t$-point FFT has much more complex hardware than $2^t$-point FFT.}, the compression ratio of each layer of \textsc{CirCNN} has to be selected among the values of $2^t$. Obviously, such restriction severely limits the potential applications of \textsc{CirCNN}.

Finally, it loses the opportunity of utilizing input sparsity. As discussed in \cite{han2016eie}\cite{albericio2016cnvlutin}, utilizing the dynamic input sparsity is an important technique to reduce the power consumption and computational time of DNN hardware. However, since \textsc{CirCNN} processes the input vector in the frequency domain, its original abundant sparsity in time domain is completely lost.
It prevents \textsc{CirCNN} from utilizing the important input sparsity for performance improvement.

\section{\textsc{PermDNN}: ALGORITHMS \& BENEFITS}
\label{sec:benefits & algo.}

\subsection{\textsc{PermDNN} Representation}

To address the drawbacks of the current 
sparsification-based and structured matrix-based
methods, we propose \textsc{PermDNN}, a new structured sparse DNN representation, 
shown in Fig. \ref{fig:bpd_matrix}(b).
Specifically, we enforce that the weight matrices of the DNN models consist of multiple permuted diagonal sub-matrices, where all the non-zero entries of the permuted diagonal matrices locate in the diagonals or permuted diagonals. In general, the $m$-by-$n$ weight matrix of one layer of \textsc{PermDNN} contains $mn/p^2$ $p$-by-$p$ permuted diagonal sub-matrices\footnote{When $m$ or $n$ is not divided by $p$, zeroes are padded.
It will not cause extra overhead since padded zeroes are not involved in computation/storage.}. As it will be shown later, this type of \textit{block-permuted diagonal weight matrix} exhibits strong spatial structure that brings significant reduction in space cost and computational cost.


\vspace{-2mm}
\subsection{Inference and Training Schemes of \textsc{PermDNN}}
\label{subsec:infer train schemes}

Based on the representation of \textsc{PermDNN}, 
we develop the computation procedures of forward and backward propagation that are the key parts for inference and training process.

In general, we assume that the weight matrix of a fully-connected layer of \textsc{PermDNN} is an $m$-by-$n$ block-permuted diagonal matrix $\mathbf{W}\in \mathbb{R}^{m\times n}$. Here $\mathbf{W}$ consists of multiple $p$-by-$p$ permuted diagonal sub-matrices. Considering there are $(m/p)\times(n/p)$ sub-matrices in total and each has its own permutation parameter, we index all the sub-matrices from 0 to $(m/p)\times(n/p)-1$ and denote $k_l$ as the value of permutation parameter for the $l$-th permuted diagonal sub-matrix. Then, for arbitrary entry $w_{ij}$ of $\mathbf{W}$, its value can be represented as:
\vspace{-2mm}
\begin{equation}
\label{eqn:wij}
w_{ij}=
\begin{cases}
  q_{k_l \times p + c} & \text{if $(c+k_l) \mod p \equiv d$}\\
  0 & \text{otherwise}
\end{cases},
\end{equation}
where $c\equiv i\mod p$, $d\equiv j\mod p$ and $l=(i/p)\times(n/p)+(j/p)$. Here $\mathbf{q}=(q_0, q_1, \dots, q_{mn/p - 1})$ is the vector that contains all the non-zero entries of $\mathbf{W}$.

\textbf{Forward Propagation for Inference.} Recall that the forward propagation during inference phase for FC layer is performed as $\mathbf{y}=\psi(\mathbf{a})=\psi(\textbf{Wx})$\footnote{Bias is combined with $\textbf{W}$ for simplicity.}, where $\psi(\cdot)$ is the activation function and $\mathbf{a}=\mathbf{Wx}=(a_1,a_2,\dots,a_m)^T$. In addition, $\mathbf{x}=(x_1,x_2,\dots,x_n)^T$ and $\mathbf{y}=(y_1,y_2,\dots,y_m)^T$ are the input and output vectors of FC layer, respectively. Accordingly, when $\mathbf{W}$ is the trained block-permuted diagonal weight matrix with entries that are represented by Eqn. (\ref{eqn:wij}), the calculation of $a_i$, as the key computation in the forward propagation, can now be simplified as $a_i = \sum_{g=0}^{n/p-1} w_{ij}x_j$, where $j\equiv (i+k_l)\mod p+gp$ and $l=g+(i/p)\times(n/p)$. Notice that here for weight storage, only the $mn/p$ -length vector $\mathbf{q}$ that contains all the non-zero entries of W needs to be stored in the inference phase. Also, the calculation of $a_i$ is significantly simplified from the original computation as $a_i = \sum_{g=0}^{n-1} w_{ij}x_j$. 




\textbf{Backward Propagation for Training.} When the FC layer is imposed with permuted diagonal structure, 
we need to ensure the FC layer of \textsc{PermDNN} always exhibits permuted diagonal structure in each iteration of training phase. Recall that the forward propagation for FC layer is performed as $\mathbf{y}=\psi(\mathbf{a})$ where $\mathbf{a}=(a_1,a_2,\dots,a_m)^T$. Then, the gradient calculation, which is the key step of the backward propagation, is performed as $\frac{\partial J}{\partial w_{ij}} = x_j \frac{\partial J}{\partial a_i}$, where $J$ is the loss function \cite{rumelhart1986learning} of the neural network. Notice that according to the principle of backpropagation \cite{rumelhart1986learning}, $x_i$ in the current layer will be backward propagated to the previous layer as $a_i$. Therefore, together with the above equation, the weight updating rule for the FC layer of \textsc{PermDNN} can be derived as:
\vspace{-1mm}
\begin{equation}
\label{eqn:upd_wij}
    w_{ij} \leftarrow w_{ij} - \epsilon x_j\frac{\partial J}{\partial a_i}, \text{for any $w_{ij}\neq 0$},
\end{equation}
\begin{equation}
\label{eqn:par_j_xj}
    \frac{\partial J}{\partial x_j} = \sum_{g=0}^{m/p-1} w_{ij}\frac{\partial J}{\partial a_i},
\end{equation}
where $i\equiv (j+p-k_l)\mod p+gp$, $l=gn/p+j/p$
and $\epsilon$ is the learning rate. 
Here in Eqn. (\ref{eqn:upd_wij}) the calculation of $\frac{\partial J}{\partial a_i}$ is aided with the computation in Eqn. (\ref{eqn:par_j_xj}) since each $x_i$ in the current layer will be backward propagated to the previous layer as $a_i$. Notice that here the weight update scheme described in Eqns. (\ref{eqn:upd_wij}) (\ref{eqn:par_j_xj}) theoretically guarantees the trained sparse network always exhibits block-permuted diagonal structure and hence provides attractive end-to-end training solution.




\vspace{-2mm}
\subsection{Extension to Convolutional Layer}
\label{subsec:conv}
\begin{figure}[!h]
\vspace{0mm}
\centering
\includegraphics[width=0.8\columnwidth]{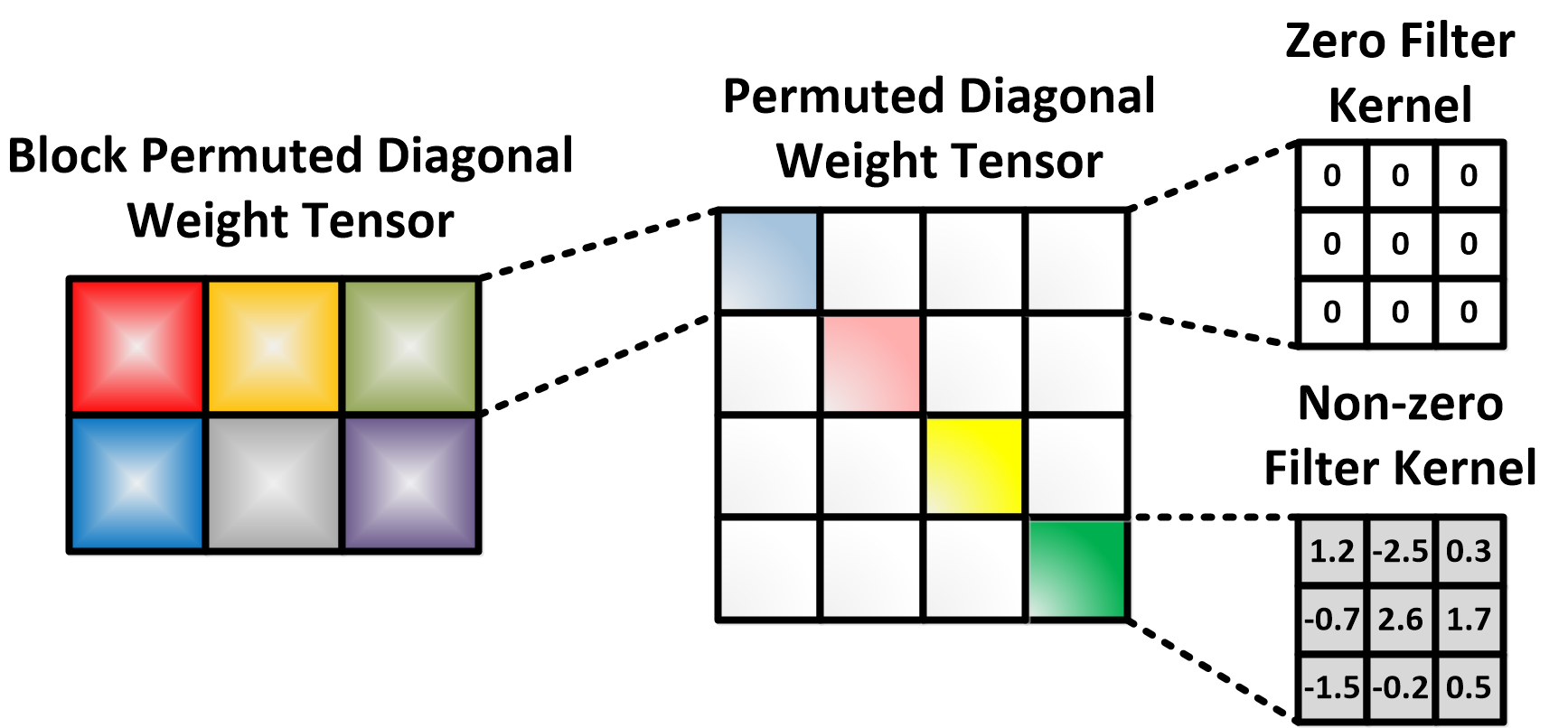}
\caption{Block-permuted diagonal weight tensor of CONV layer.}
\vspace{0mm}
\label{fig:weight_tensor}
\end{figure}

\textbf{Forward Propagation for Inference on CONV Layer.} The idea of imposing permuted diagonal structure on the weight matrix of FC layer can be further generalized and applied to the weight tensor of CONV layer. Recall that the weight tensor of a CONV layer is a 4D tensor that can be viewed as a "macro" matrix with each entry being a filter kernel; therefore, as illustrated in Fig. \ref{fig:weight_tensor}, the permuted diagonal structure can be imposed on the input channel and output channel dimensions of the weight tensor. Then, similar to the case for FC layer, the forward propagation for inference on CONV layer using permuted diagonal matrix can be described as follows:
\vspace{-6mm}

\begin{equation}
\vspace{-1mm}
\label{eqn:conv_forward}
    \mathcal{Y}(i,x,y)=\sum_{g=0}^{c_2/p-1}\sum_{w=0}^{w_1-1}\sum_{h=0}^{h_1-1}\mathcal{F}(i,j,w,h)\mathcal{X}(j,x-w,y-h),
\end{equation}
\vspace{-1mm}
where $\mathcal{X}\in\mathbb{R}^{c_0\times w_0\times h_0}$, $\mathcal{Y}\in\mathbb{R}^{c_2\times w_2\times h_2}$, $\mathcal{F}\in\mathbb{R}^{c_0\times c_2\times w_1\times h_1}$ represent the input, output and weight tensors of convolutional layer, respectively. Here $w_i$ and $h_i$ for $i$=0,1,2 are the width and height of the input, kernel and output tensor, respectively. $c_0$ and $c_2$ are the numbers of input channels and output channels. In addition, $j\equiv(i+k_l)\mod p+gp$ and $l=g+(i/p)\times(n/p)$.

\textbf{Backward Propagation for Training on CONV Layer.} Similar to the FC layer, in order to ensure the convolutional layer of \textsc{PermDNN} always exhibits the permuted diagonal structure during the training phase, the corresponding weight updating procedure needs to be re-designed. Following the similar steps of deriving training procedure for FC layer, the weight update rule in the back propagation on CONV layer can be described as follows:
\vspace{-7mm}

\begin{align}
\label{eqn:conv_backward_F}
\begin{split}
    \mathcal{F}(i,j,w,h) \leftarrow& 
    \mathcal{F}(i,j,w,h)-\epsilon\sum_{x=0}^{w_2}\sum_{y=0}^{h_2}\mathcal{X}(i,x-w,y-h)\\
    &\times \frac{\partial J}{\partial \mathcal{Y}(i,x,y)}, \text{for any $\mathcal{F}(i,j,w,h)\neq 0$},
\end{split}
\end{align}

\vspace{-9mm}

\begingroup
\thickmuskip=0mu
\begin{equation}
\label{eqn:conv_backward_X}
\frac{\partial J}{\partial \mathcal{X}(i,j,x)}{=}
\sum_{g=0}^{c_0{/}p{-}1}\sum_{w=0}^{w_1{-}1}\sum_{h=0}^{h_1{-}1}\mathcal{F}(i,j,w,h)  \frac{\partial J}{\partial \mathcal{Y}(i,x{+}w,y{+}h)},
\end{equation}
\endgroup

where $i=(j+p-k_l) \mod p+gp$ and $l=gn/p+j/p$. Notice that here in Eqn. \ref{eqn:conv_backward_F} the calculation of $\frac{\partial J}{\partial \mathcal{Y}(i,x,y)}$ is aided with the computation in Eqn. \ref{eqn:conv_backward_X} since each $\mathcal{X}(j,x,y)$ in the current layer will be backward propagated to the previous layer as $\mathcal{Y}(j,x,y)$. Again, the weight update scheme described in Eqns. (\ref{eqn:conv_backward_F}) (\ref{eqn:conv_backward_X}) theoretically guarantees the trained sparse network always exhibits block-permuted diagonal structure.

\vspace{-2mm}
\subsection{Test Accuracy and Compression Ratio}
\label{subsec:accuracy}

By leveraging the forward and backward propagation schemes in Eqns. (\ref{eqn:upd_wij})-(\ref{eqn:conv_backward_X}), the \textsc{PermDNN} models can be trained from scratch and tested. Table \ref{table:alexnet} - Table \ref{table:wrn} show the task performance and compression ratio of different \textsc{PermDNN} models on different types of datasets. Notice that for one FC/CONV layer with block size $p$ for its permuted diagonal (PD) weight matrix/tensor, the compression ratio for that layer is $p$. The details of the experimental setting are described as follows:

    \begin{itemize}
        \item \textbf{AlexNet \cite{krizhevsky2012imagenet}:} The block sizes ($p$) for the permuted diagonal weight matrices of three FC layers (FC6, FC7 and FC8) are set as different values (10, 10 and 4) for different FC layers.
        \item \textbf{Stanford Neural Machine Translation (NMT) \cite{luong2015stanford}:} This is a stacked LSTM model containing 4 LSTMs with 8 FC weight matrices for each LSTM. In the experiment the value of $p$ for all the FC layers is set as 8. 
        \item \textbf{ResNet-20 \cite{he2016deep}:} In this experiment for the group of CONV layers without 1x1 filter kernel, the value of $p$ is set as 2. For the group of CONV layers with 1x1 filter kernel, $p$ is set as 1.
        \item \textbf{Wide ResNet-48 \cite{zagoruyko2016wide}:} The widening parameter of this model is 8. For the group of CONV layers without 1x1 filter kernel, the value of $p$ is set as 4. For the group of CONV layers with 1x1 filter kernel, $p$ is set as 1.
        \item \textbf{Selection of Permutation value ($k_l$):} $k_l$ can be selected via either natural indexing or random indexing. Our simulation results show no difference between task performance for these two setting methods. Table \ref{table:alexnet} - Table \ref{table:wrn} are based on natural indexing. For instance, for a 4-by-16 block-permuted diagonal weight matrix with $p=4$,  $k_0 \sim k_3$ is set as $0 \sim 3$.
    \end{itemize}


Table \ref{table:alexnet} - Table \ref{table:wrn} show that imposing permuted diagonal structure to DNN models enables significant reduction in the weight storage requirement for FC or CONV layers. Meanwhile, the corresponding task performance, in terms of test accuracy (for computer vision) or BLEU scores \cite{papineni2002bleu} (for language translation), are still retained as the same or only exhibits negligible degradation. In short, \textsc{PermDNN} models can achieve high compression ratios in network size and strong spatial network structure, and simultaneously, preserve high task performance.
It makes the corresponding hardware-friendly architecture (Section~\ref{sec:architecture}) very attractive.
\vspace{-1mm}

\begin{table}[!h]
  \caption{AlexNet on ImageNet \cite{deng2009imagenet}. PD: Permuted Diagonal.}
  \vspace{0mm}
  \centering
  \begin{tabular}{|>{}c|>{}c|>{}c|>{}c|}
    \hline
     AlexNet & \specialcell{Block size ($p$) for \\ PD weight matrix \\ of FC6-FC7-FC8} & \specialcell{Top-5 \\ Acc.} &  \specialcell{Compression \\ for overall \\ FC layers} \\
     \hline
     Original 32-bit float & 1-1-1 & 80.20\% & 234.5MB(1$\times$) \\
     \hline
     32-bit float with PD & 10-10-4 & 80.00\% & 25.9MB(9.0$\times$) \\
     \hline
     16-bit fixed with PD & 10-10-4 & 79.90\% & 12.9MB(18.1$\times$) \\
     \hline
    \end{tabular}
  \label{table:alexnet}
\end{table}
\vspace{-5mm}

\begin{table}[!h]
  \caption{Stanford NMT (32-FC layer LSTMs) on IWSLT15 for English-Vietnamese Translation.}
  \vspace{0mm}
  \centering
  \begin{tabular}{|>{}c|>{}c|>{}c|>{}c|}
    \hline
    \specialcell{Stanford NMT \\(32-FC layer LSTMs)\footnote{Here one FC in LSTM means one component weight matrix.}} & \specialcell{Block size ($p$) for \\ PD weight matrix \\ of ALL FC Layers} & \specialcell{BLEU \\ Points} & \specialcell{Compression \\ for overall \\ FC layers} \\
     \hline
     Original 32-bit float & 1 & 23.3 & 419.4MB(1$\times$) \\
     \hline
     32-bit float with PD & 8 & 23.3 & 52.4MB(8$\times$) \\
     \hline
     16-bit fixed with PD & 8 & 23.2 & 26.2MB(16$\times$) \\
     \hline
    \end{tabular}
  \vspace{-5mm}
  \label{table:nmt}
\end{table}

\begin{table}[!h]
  \caption{ResNet-20 on CIFAR-10\cite{krizhevsky2009learning}.}
  \centering
  \begin{tabular}{|>{}c|>{}c|>{}c|>{}c|}
    \hline
     ResNet-20 & \specialcell{Block size ($p$) for \\ PD weight tensor \\ of CONV layers} & Acc. & \specialcell{Compression \\ for overall \\ CONV Layers} \\
     \hline
     Original 32-bit float & 1 & 91.25\% & 1.09MB(1$\times$)\\
     \hline
     32-bit float with PD & 2 for most layers & 90.85\% & 0.70MB(1.55$\times$)\\
     \hline
     16-bit fixed with PD & 2 for most layers& 90.6\% & 0.35MB(3.10$\times$) \\
     \hline
    \end{tabular}
  \vspace{-5mm}
  \label{table:resnet}
\end{table}

\begin{table}[!h]
  \caption{Wide ResNet-48 on CIFAR-10.}
  \vspace{0mm}
  \centering
  \begin{tabular}{|>{}c|>{}c|>{}c|>{}c|}
    \hline
     Wide ResNet-48 & \specialcell{Block size ($p$) for \\ PD weight tensor \\ of CONV layers} & Acc. & \specialcell{Compression \\ for overall \\ CONV layers}\\
     \hline
     Original 32-bit float & 1 & 95.14\% & 190.2MB(1$\times$) \\
     \hline
     32-bit float with PD & 4 for most layers& 94.92\% & 61.9MB(3.07$\times$) \\
     \hline
     16-bit fixed with PD & 4 for most layers & 94.76\% & 30.9MB(6.14$\times$) \\
     \hline
    \end{tabular}
  \label{table:wrn}
\end{table}
\vspace{0mm}

\subsection{Outline of Theoretical Proof on Universal Approximation}
In \textsc{CirCNN} the \textit{universal approximation property} of block-circulant matrix-based neural network was given to theoretically prove the effectiveness of using circulant matrices. In this work we discover that the \textsc{PermDNN} also exhibits the universal approximation property, thereby making the rigorous foundation for our proposed permuted diagonal structure-imposing method. The details of the proof will be provided in an individual technical report and this subsection gives a brief outline of the proof as follows.
First, we prove that the "connectedness" of \textsc{PermDNN}, -- that means, thanks to the unique permuted diagonal structure of each block, when $k_l$ is not identical for all permuted diagonal matrices, the sparse connections between adjacent block-permuted diagonal layers do not block away information from any neuron in the previous layer. Based on this interesting property, we further prove that the function space achieved by the block-permuted diagonal networks is dense. Finally, with the Hahn-Banach Theorem we prove that there always exists a block-permuted diagonal neural network that can closely approximate any target continuous function defined on a compact region with any small approximation error, thereby showing the universal approximation property of block-permuted diagonal networks. Besides, we also derive that the error bound of this approximation error is in the order of $O(1/n)$, where $n$ is the number of model parameters. Consequently, the existence of universal approximation property of \textsc{PermDNN} theoretically guarantees its effectiveness on different DNN types and applications.

\vspace{0mm}
\subsection{Applicability on the Pre-trained Model}
\begin{figure}[!h]
\vspace{0mm}
\centering
\includegraphics[width=1\columnwidth]{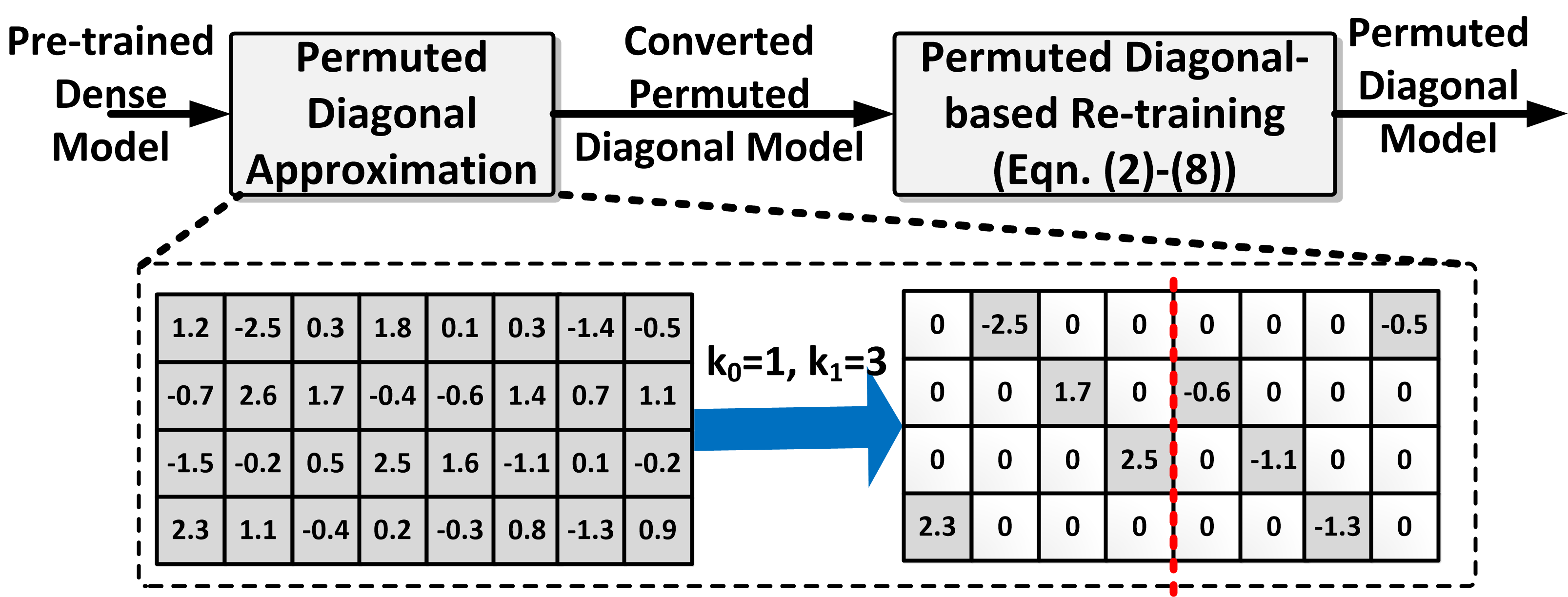}
\caption{Train a \textsc{PermDNN} from a pre-trained dense model.}
\vspace{0mm}
\label{fig:approx}
\end{figure}

Besides training from scratch, the permuted diagonal matrix-based network model can also be obtained from a pre-trained dense model. Fig. \ref{fig:approx} illustrates the corresponding procedure, which consists of two steps: permuted diagonal approximation and re-training/fine-tuning. First, the original dense weight matrices/tensors need to be converted to permuted diagonal matrices/tensors via \textit{permuted diagonal approximation}. The mechanism of permuted diagonal approximation is to convert a non-permuted diagonal matrix/tensor to a permuted diagonal format by only keeping the entries in the desired permuted diagonal positions. \textbf{Mathematically, such approximation is the optimal approximation in term of $l_2$ norm measurement on the approximation error}. After that, the converted model already exhibits permuted diagonal structure and then can be further re-trained/fine-tuned by using Eqns. (\ref{eqn:upd_wij})-(\ref{eqn:conv_backward_X}) to finally obtain a high-accuracy permuted diagonal network model. Such two-step generating approach is applied to all types of pre-trained models and can lead to high accuracy. For instance, for pre-trained dense LeNet-5 model on MNIST dataset, with $p=4$ for CONV layer and $p=100$ for FC layer, the finally converted permuted-diagonal network after re-training achieves 99.06\% test accuracy and overall $40\times$ compression ratio without using quantization.

\subsection{\textsc{PermDNN} vs Unstructured Sparse DNN}
\label{subsec:benefits of pd}

\begin{figure}[!h]
\vspace{-3mm}
\centering
\includegraphics[width=1\columnwidth]{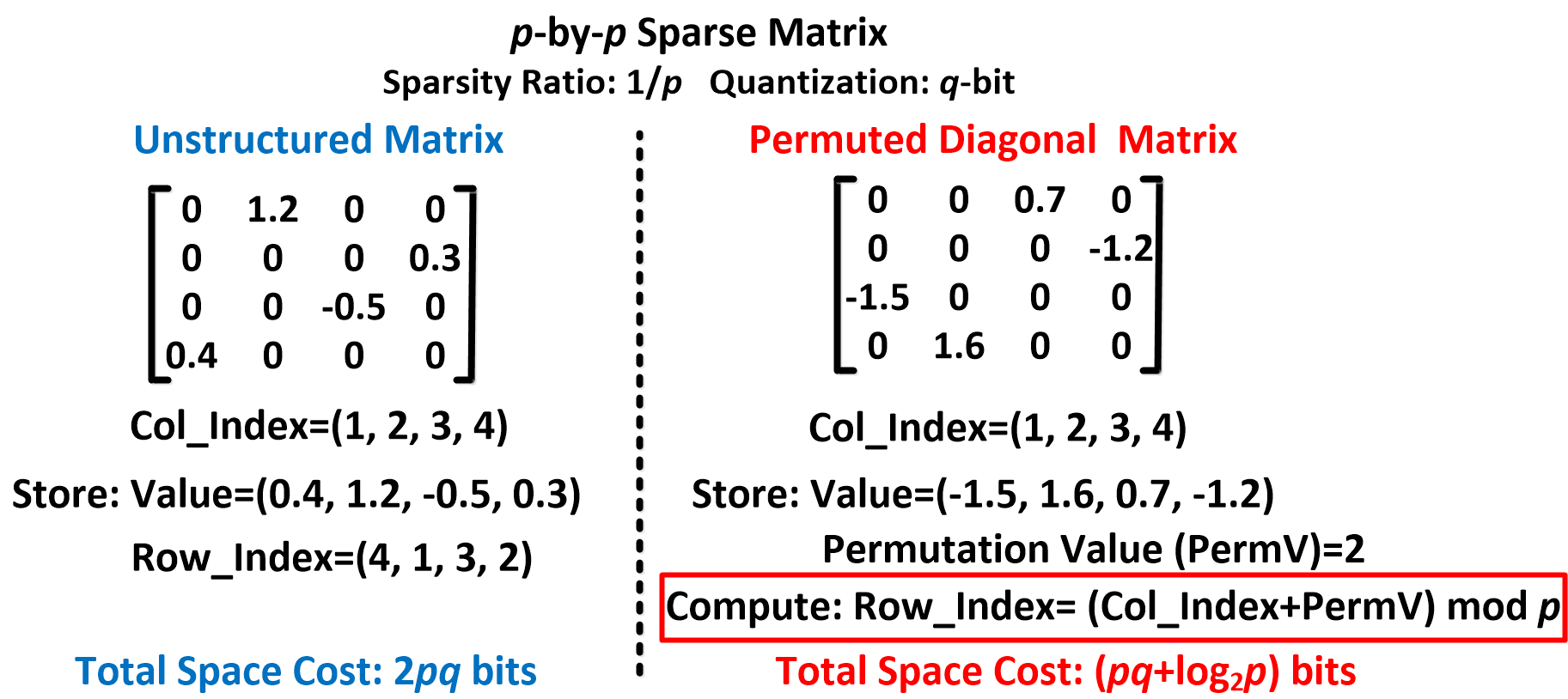}
\caption{Storage requirement comparison.}
\vspace{-5mm}
\label{fig:matrix_comp}
\end{figure}

Compared to the existing network sparsification approaches, the proposed \textsc{PermDNN} enjoys several attractive advantages:

First, \textit{\textsc{PermDNN} is a hardware-friendly model}. As illustrated in Fig. \ref{fig:matrix_comp}, due to the inherent regular structure of permuted diagonal matrix, the position of each non-zero entry can now be calculated using very simple modulo operation\footnote{As shown in Fig. \ref{fig:acc_sel}, modulo circuit is simple by using LSB bits.}, thereby completely eliminating the needs of storing the indices of entries. From the perspective of hardware design, this elimination means that the permuted diagonal matrix-based \textsc{PermDNN} completely avoids the huge space/computation overhead incurred by the complicated weight indexing/addressing in the state-of-the-art sparse DNN accelerator (e.g. EIE), and hence achieves significant reduction in the space requirement and computational cost for DNN implementations. 

Second, \textit{\textsc{PermDNN} provides controllable and adjustable compression and acceleration schemes.} The reductions in the model sizes and number of arithmetic operations are no longer heuristic based and unpredictable. Instead it can now be precisely and deterministically controlled by adjusting the value of $p$. This further provides great benefits to explore the design space exploring the tradeoff between hardware performance and test accuracy.

Finally, \textit{\textsc{PermDNN} enables direct end-to-end training while preserving high accuracy}. Since the structure of sparse weight matrices of \textsc{PermDNN} can now be pre-determined by the model designers at the initialization stage of training, with our training algorithm that
preserves this fixed structure, the entire structured sparse network can be trained from scratch, completely avoiding the increasing complexity incurred by the extra iterative pruning and/or re-training process in the conventional unstructured sparse DNN training schemes. 

\subsection{\textsc{PermDNN} vs \textsc{CirDNN}}
\label{subsec:permdnn vs cirdnn}

While \textsc{PermDNN} and \textsc{CirDNN}
are both based on the structured matrices,
\textsc{PermDNN} does not suffer from the drawbacks
of \textsc{CirDNN}.
Table~\ref{table:summary circnn and permdnn}
compares the two methods. 


Most importantly, \textsc{PermDNN} uses simpler arithmetic computation. Different from complex number computation-based \textsc{CirDNN}, the computation of \textsc{PermDNN} is purely based on real numbers, thereby requiring significantly lower arithmetic cost than \textsc{CirDNN} with the same workload and same compression ratio. 

Second, \textsc{PermDNN} allows compression ratio to be flexibly adjusted. Different from \textsc{CirDNN}, the computation in \textsc{PermDNN} does not have any restrictions on the size of component permuted diagonal matrix. Therefore, the hardware architecture of \textsc{PermDNN} can support different compression ratios based on the 
needs. This makes \textsc{PermDNN} more attractive for a wide range of applications.

Finally, the fundamental distinction is that \textsc{PermDNN} can fully utilize input sparsity. Since the computation of \textsc{PermDNN} is in the time domain, the important input sparsity can still be leveraged by \textsc{PermDNN} to further reduce computational cost and power consumption. Because sparsity of input vectors widely exists in numerous applications, this benefit greatly expands the advantage of \textsc{PermDNN} over \textsc{CirDNN}.
\vspace{0mm}

\begin{table}[h]
  \caption{Advantages of \textsc{PermDNN} over \textsc{CirCNN}.}
  \centering
  \begin{tabular}{|c|c|c|c|}
    \hline
      & \textsc{CirCNN} & \textbf{\textsc{PermDNN}} \\ \hline
    \textbf{Arithmetic Operation} & Complex number-based & Real number-based \\ \hline
    \textbf{Flexible Compression} & No & Yes \\
    \hline
    \textbf{Utilize Input Sparsity} & No & Yes \\
    \hline
  \end{tabular}
  \vspace{0mm}
  \label{table:summary circnn and permdnn}
\end{table}

\section{\textsc{PermDNN}: ARCHITECTURE}
\label{sec:architecture}

In this section, we develop the
\textsc{PermDNN} architecture based on the 
proposed structure and algorithms.
Similar to most existing DNN accelerators \cite{chen2016eyeriss}\cite{parashar2017scnn}\cite{han2016eie}, the \textsc{PermDNN} architecture is
designed for inference tasks.

\subsection{Data Mapping and Processing Scheme}
\label{subsec:data mapping}
\begin{figure*}[!h]
\centering
\includegraphics[width=0.7\textwidth]{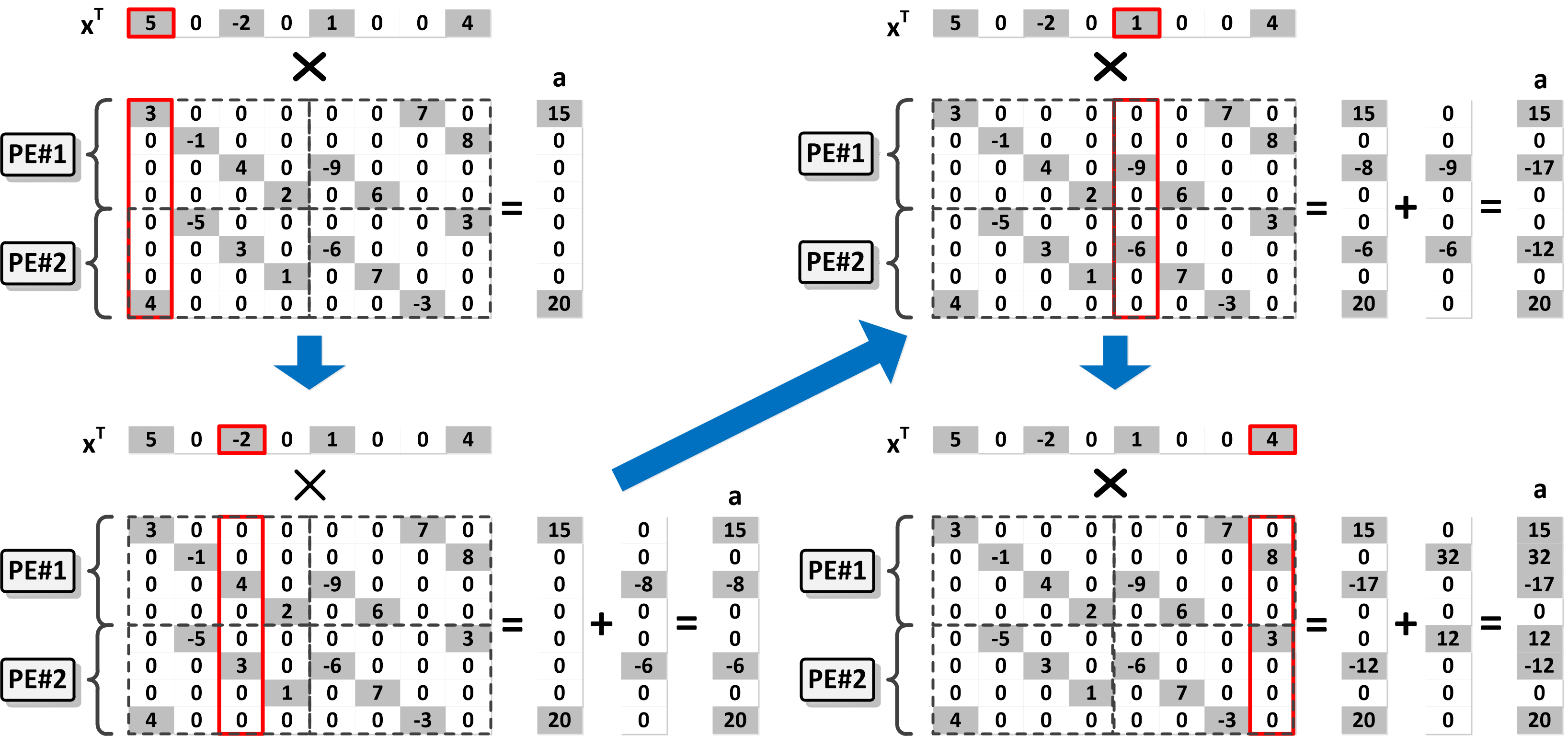}
\caption{Example column-wise processing procedure with input zero-skipping scheme.}
\vspace{-5mm}
\label{fig:data_mapping}
\end{figure*}

In general, the proposed computing engine is a scalable array of processing elements (PEs). Each PE performs the forward propagation on part of the weight matrix.
Different PEs perform independent and parallel computations to maximize the processing throughput.
Fig. \ref{fig:data_mapping} illustrates this partition scheme on an 8-by-8 block-permuted diagonal weight matrix with $p=4$. We can see that the 4-by-4 permuted diagonal sub-weight matrices belonging to the same block row are processed by the same PE\footnote{In general, each PE can be in charge of multiple permuted diagonal matrices along column direction.}. To support such arrangement, 
the non-zero entries of these sub-matrices are stored in each PE's associated SRAM. Therefore, the entire weight matrix of the FC layer is stored in a distributed manner across multiple SRAM banks to enable parallel processing of multiple PEs. 

Besides, in order to fully utilize the potential dynamic sparsity in the input vector $\mathbf{x}$ 
(also the activation vector output $\mathbf{y}$ from previous layer), the entire \textsc{PermDNN} adopts \textit{column-wise} processing style. Specifically, as illustrated in Fig. \ref{fig:data_mapping}, in each clock cycle all PEs perform the multiplications between $x_i$, the non-zero entry of $\mathbf{x}$, and $\mathbf{w_i}$, the corresponding column vector of weight matrix $\mathbf{W}$. The products are then accumulated for updating the intermediate values of the corresponding entries in the output vector $\mathbf{a}$. After all the non-zero entries of current $\mathbf{w_i}$ have been multiplied with $x_i$, the PEs will then move on to perform the multiplications between $x_j$ and $\mathbf{w_j}$, where $x_j$ is the next non-zero entry that follows $x_i$ in the $\mathbf{x}$. After all the $x_j$ and $\mathbf{w_j}$ have been processed, the final calculated values of all the entries of output vector $\mathbf{a}$ are available simultaneously. Obviously, by adopting this column-wise processing scheme, the computations that are involved with zero entries of $\mathbf{x}$ can be completely skipped, thereby leading to significant saving in energy consumption when there exists non-negligible sparsity in $\mathbf{x}$.

\subsection{Overall Architecture}
\label{subsec:overall arch}
\begin{figure}[!h]
\centering
\includegraphics[width=0.9\columnwidth]{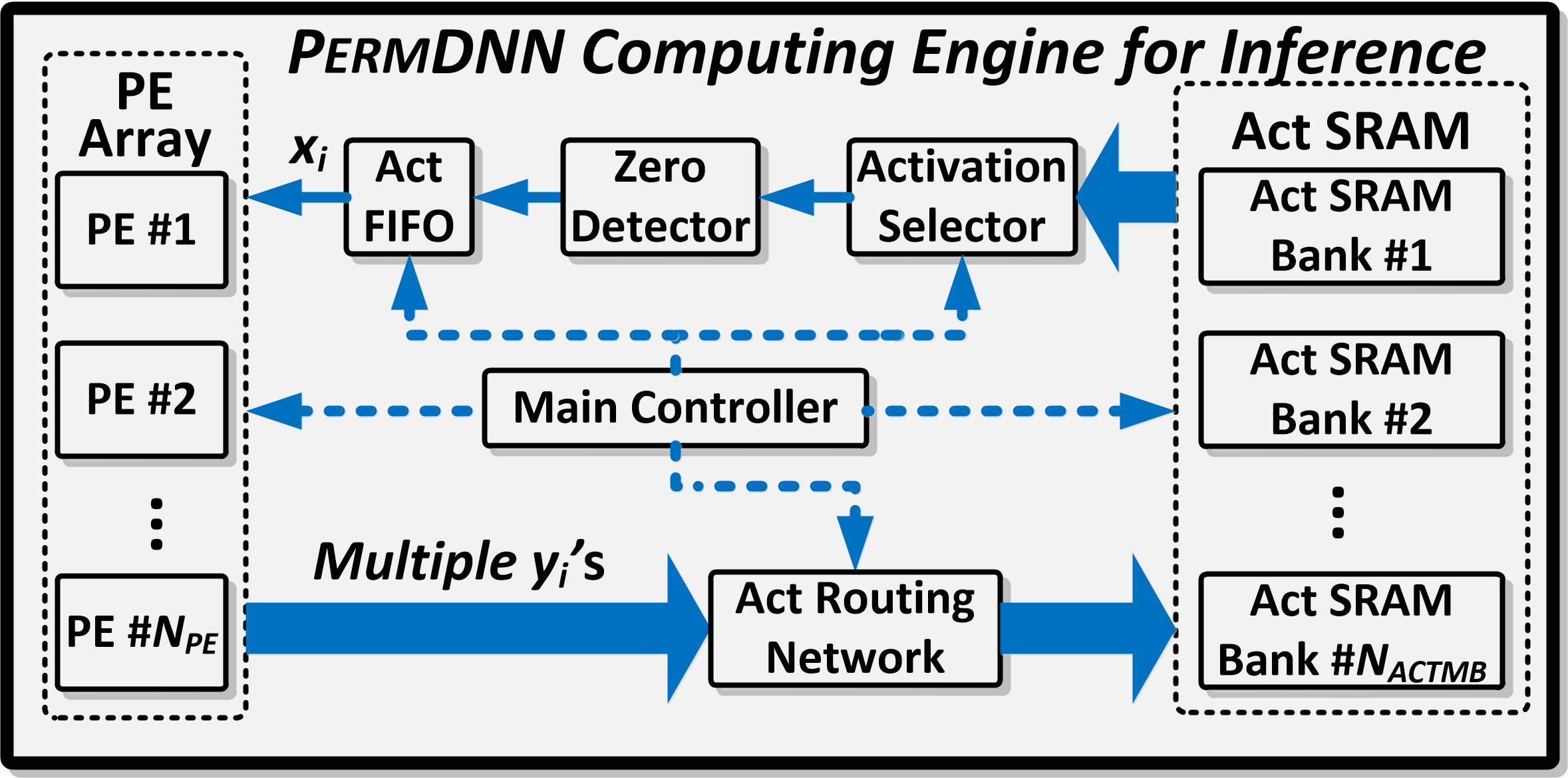}
\caption{Overall architecture of \textsc{PermDNN} hardware.}
\vspace{0mm}
\label{fig:overall_arch}
\end{figure}

Based on the data mapping and processing scheme, 
the overall architecture of \textsc{PermDNN} computing engine is shown in Fig. \ref{fig:overall_arch}. The entire design consists of an array of $N_{PE}$ PEs that perform the kernel $M\times V$ operations and the non-linear activation operations. After all the $y_i$'s, 
the entries of output activation vector $\mathbf{y}$ for the current FC layer, have been calculated and stored in the PEs, they are then written to the activation SRAM. The writing operation is performed in a group-writing manner: the entire activation SRAM is partitioned into $N_{ACTMB}$ banks, where each SRAM bank is in charge of the $y_i$'s from $N_{ACC}/N_{PE}$ PEs. In each clock cycle, among all the PEs that belong to the same SRAM bank, one of them outputs $W_{ACTM}/q$ activation values $y_i$'s to its corresponding SRAM bank, where $W_{ACTM}$ and $q$ are the width of activation SRAM and the bit-width of $y_i$, respectively. Consider there are in total $N_{ACTMB}W_{ACTM}/q$ $y_i$'s that are simultaneously written to the activation SRAM in one cycle. An activation routing network is designed to ensure each $y_i$ is correctly written to the target position in the activation SRAM.

In the reading phase of activation SRAM, as described in Section \ref{subsec:data mapping}, each time only one non-zero activation value $x_i$ is fetched and broadcasted to all the PEs. To achieve that, with the help of control signals from main controller, an activation selector is designed to select the correct $x_i$ from multiple activation SRAM banks. After the examination from a zero-detector, the non-zero $x_i$ is then sent to an activation FIFO for its broadcast to PE arrays. The purpose of using this activation FIFO is to build up a backlog for the non-zero $x_i$'s, thereby ensuring that the PEs can always receive their required $x_i$ for the current computation in time.

\subsection{Processing Element}
\label{subsec:PE}
\begin{figure}[!h]
\centering
\vspace{0mm}
\includegraphics[width=0.9\columnwidth]{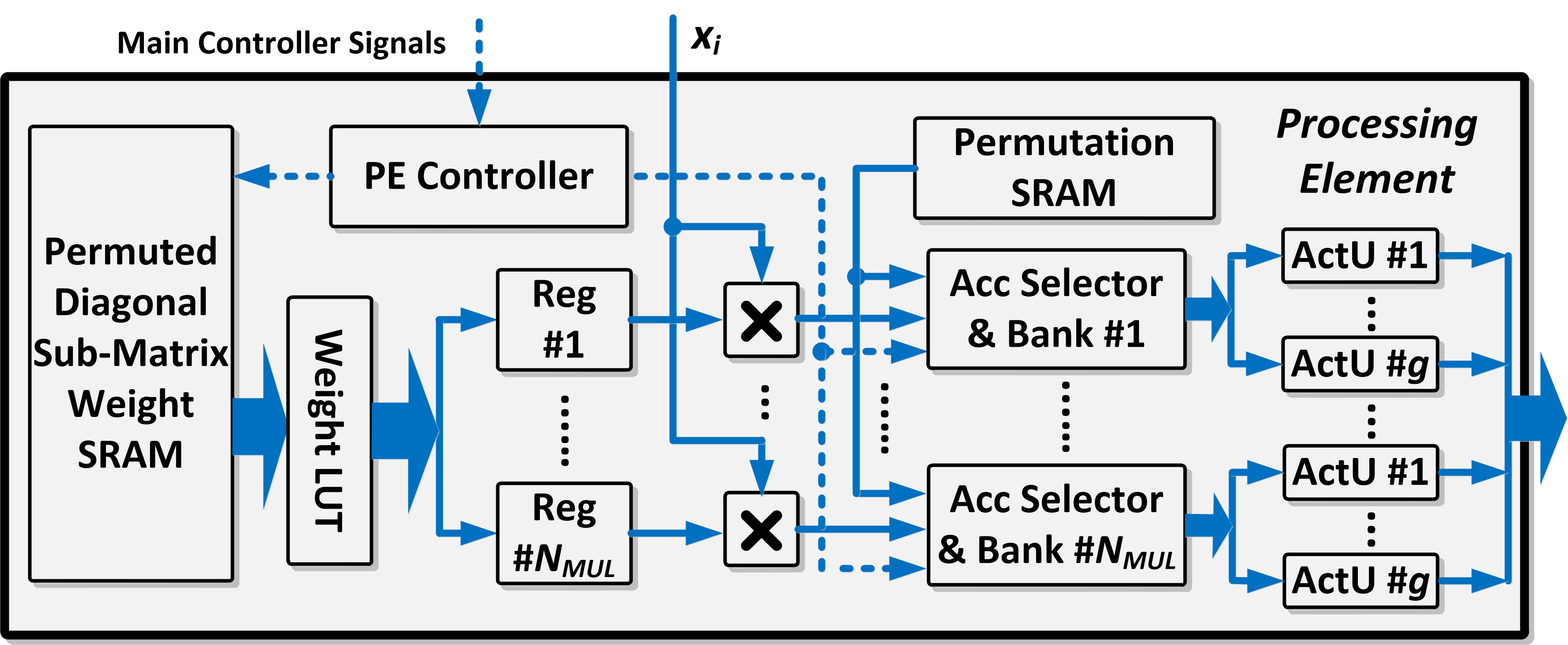}
\caption{Inner architecture of PE.}
\vspace{0mm}
\label{fig:pe}
\end{figure}

Fig. \ref{fig:pe} shows the inner architecture of the PE. Here, each PE is equipped with its own SRAM that stores the non-zero entries of part of block-permuted diagonal weight matrices. Similar to EIE, a weight lookup table (LUT) is located in the PE to support weight sharing strategy \cite{han2015learning} if needed. In that case, the weight SRAM of the PE stores the virtual weight tag (or so-called clustered class) that each actual weight has. After being decoded by the weight LUT, $N_{MUL}$ actual weights are multiplied with the input $x_i$ and the products are sent to a group of accumulators in each clock cycle. Because each PE is equipped with $N_{ACC}$ accumulators, which corresponds to $N_{ACC}$ rows of weight matrix that one PE is in charge of, $N_{MUL}$ accumulation selectors are designed to ensure those products are routed to the target accumulators. Specifically, as it will be elaborated in detail later, the permutation parameters of the corresponding permuted diagonal weight sub-matrices are used to enable very low-cost selection. For each selector, it is associated with its own accumulator bank, and each accumulator bank contains $g=N_{ACC}/N_{MUL}$ accumulators that always contain a target accumulator for the input product. After all PEs finish their processing for the current column of weight matrix, the calculated $a_i$'s in the accumulators are sent to the activation units (ActU) to produce $y_i$'s. Here, each activation unit can be reconfigured to act as either Rectified Linear Unit (ReLU) or hypertangent function ($tanh(\cdot)$) unit to support the needs of different applications.

\textbf{Weight SRAM.} To reduce the read/write energy consumption for the weight SRAM, we further break the weight SRAM bank of each PE into multiple weight SRAM sub-banks. By adopting this strategy, in each cycle only one weight SRAM sub-bank is selected while the others are disabled to save energy. Accordingly, in order to achieve the full utilization of all the multipliers in most cases, the width of all the weight SRAM sub-banks is selected to ensure that each row of every SRAM sub-bank at least contains $N_{MUL}$ weight entries.

\begin{figure}[!h]
\centering
\includegraphics[width=1\columnwidth]{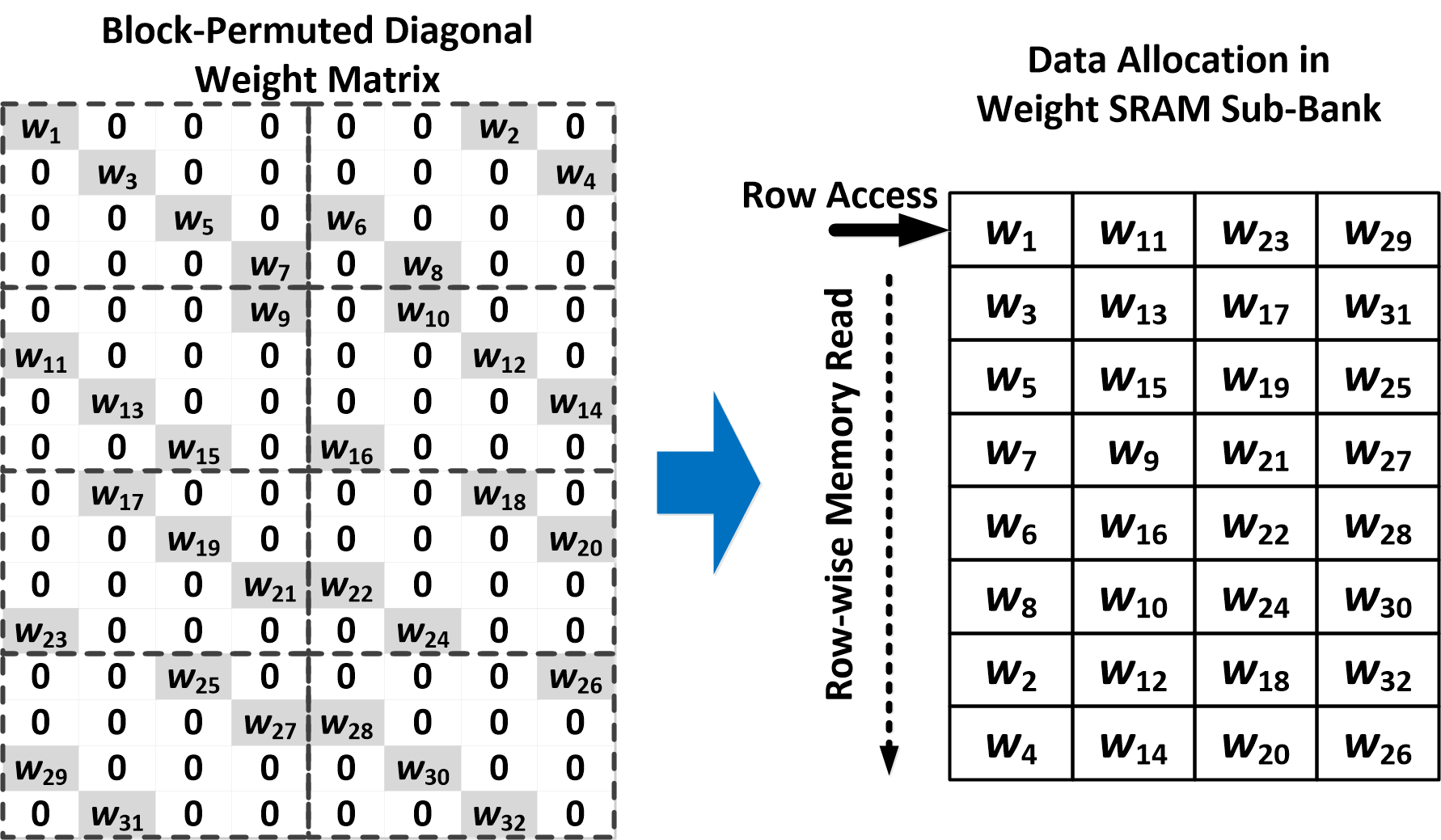}
\caption{Data allocation in weight SRAM}
\vspace{0mm}
\label{fig:memory_layout}
\end{figure}

Besides, to accommodate the column-wise processing scheme and the block-permuted diagonal structure of weight matrix, the non-zero weight is specially arranged in the weight SRAM of each PE. Fig. \ref{fig:memory_layout} illustrates the data allocation in one weight SRAM sub-bank for one block-permuted diagonal weight matrix. Here, each row of the weight SRAM sub-bank stores the non-zero entries of the same column of the weight matrix. By adopting this transpose-like data layout, each row access to the weight SRAM sub-bank can fetch the required data for column-wise data processing. 
Because each column of one permuted diagonal sub-matrix only has one non-zero entry, each PE always receives and processes the same number of non-zero weights, thereby completely eliminating the risk of load imbalance that prior work \cite{han2016eie} may suffer from.
\vspace{0mm}

\begin{figure}[!h]
\centering
\includegraphics[width=0.9\columnwidth]{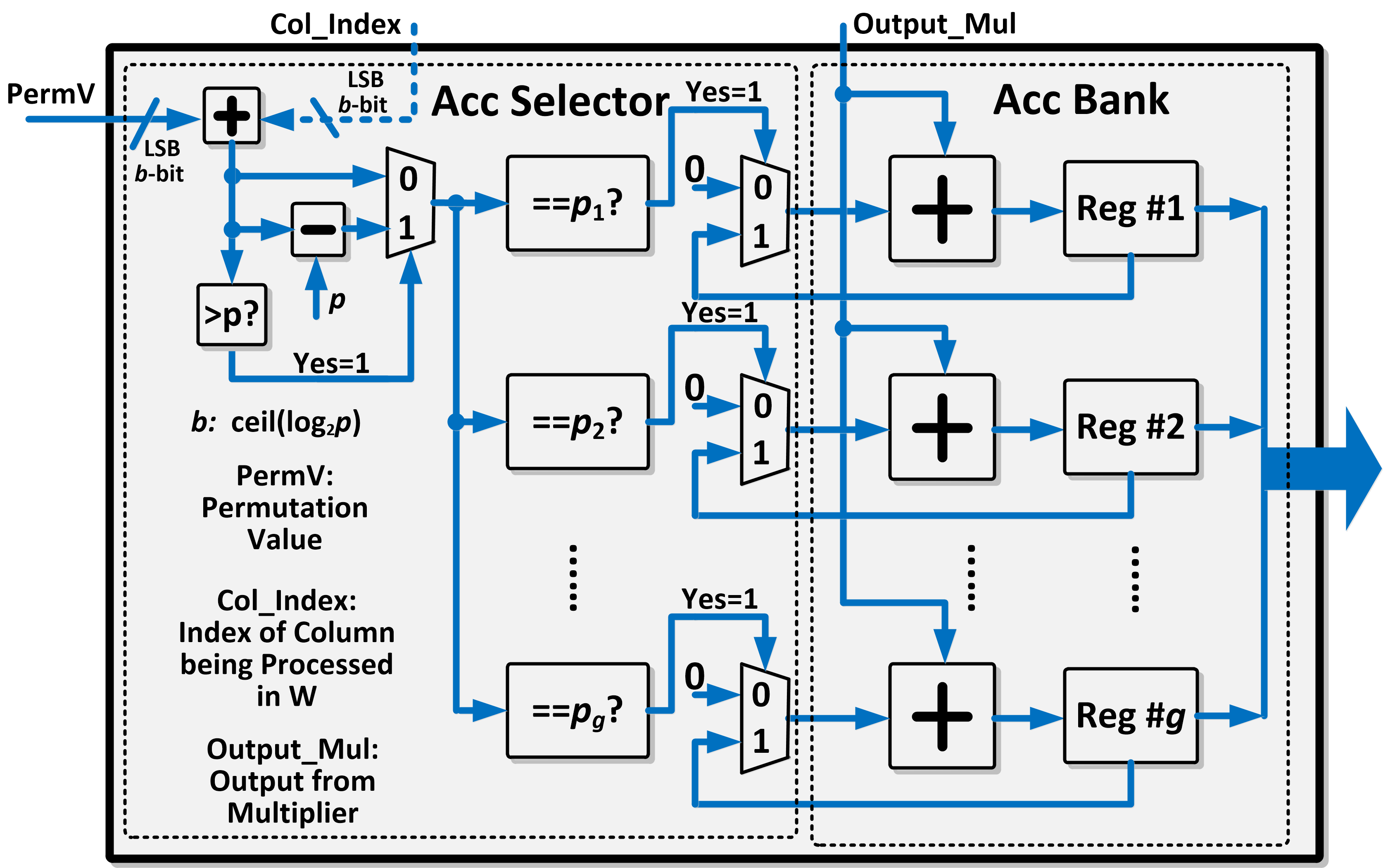}
\caption{Inner architecture of accumulation selector and bank.}
\vspace{0mm}
\label{fig:acc_sel}
\end{figure}

\textbf{Accumulation Selector \& Bank.} 
Since the number of accumulators ($N_{ACC}$) is typically much larger than that of multipliers ($N_{MUL}$) in each PE, accumulation selectors are required to route the outputs of multipliers to the target accumulators. As shown in Fig. \ref{fig:acc_sel}, an accumulation selector consists of two parts, an index calculator and an array of reconfigurable comparators and multiplexers. The index calculator is used to calculate the correct row index of the non-zero entry in the permuted diagonal sub-matrix that the current PE is in charge of. Specifically, due to the unique structure of permuted diagonal matrix, such calculation is essentially the modulo operation between the sum of permutation value and column index and the size of permuted diagonal matrix $p$; and hence it can be easily implemented using a circuit consisting of $b$-width adder, subtractor and comparator, where $b=ceil(\log_2p)$ and $ceil(\cdot)$ is the ceiling function. After the row index is calculated, among the bank of accumulators, the target accumulator is first identified by the array of comparators and multiplexers, and then it is updated using the output of the multiplier.
\vspace{0mm}\

\textbf{Permutation SRAM.} The permutation values (PermVs) sent to accumulation selectors are provided by permutation SRAM. Different from weight SRAM and activation SRAM, in each PE the permutation SRAM consists of a single bank instead of multiple partitions. In addition, each row of permutation SRAM contains multiple permutation values to support the computation in different accumulation selectors. Given that the bit-width of permutation value is typically small (as $\log_2p$), the width of permutation SRAM is not very large.

\subsection{Design for Flexibility}
\label{subsec:design for flex}
\vspace{0mm}
Because the value of $p$ and network size vary with models, the PE of \textsc{PermDNN} is designed to provide flexibility for different needs. In general, designing a flexible PE involves the
considerations of several factors: $N_{ACC}$, the number of accumulators in each PE; $N_{MUL}$, the number of multipliers in each PE; $N_{ROWPE}=m/N_{PE}$, the number of rows that each PE is in charge of; and $p$. 
In the following, we describe the computation scheme of PE in three cases.

\begin{figure*}[!t]
\centering
\begin{subfigure}{\textwidth}
{\includegraphics[width=0.9\textwidth]{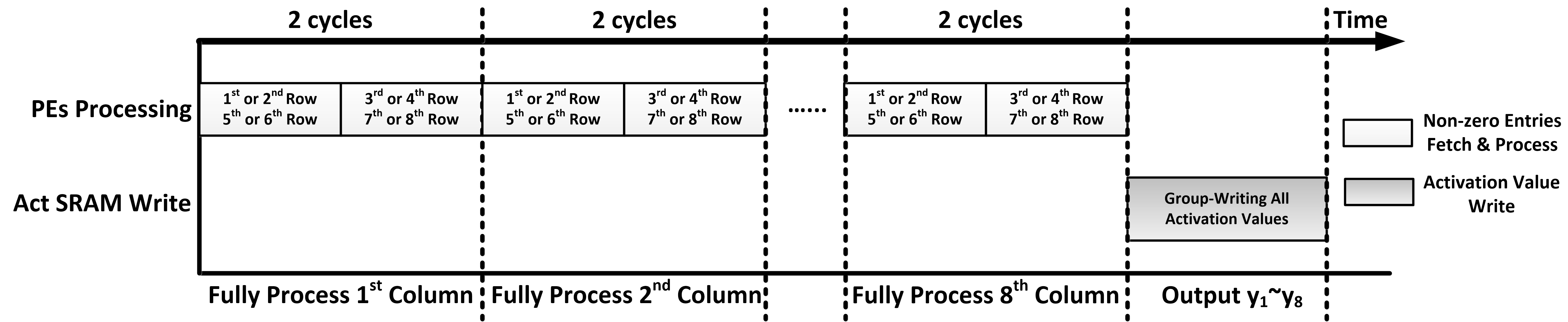}%
\label{fig_first_case}
\caption{Computation scheme with $p=2$.}
}
\end{subfigure}
\\
\begin{subfigure}{\textwidth}
\includegraphics[width=0.9\textwidth]{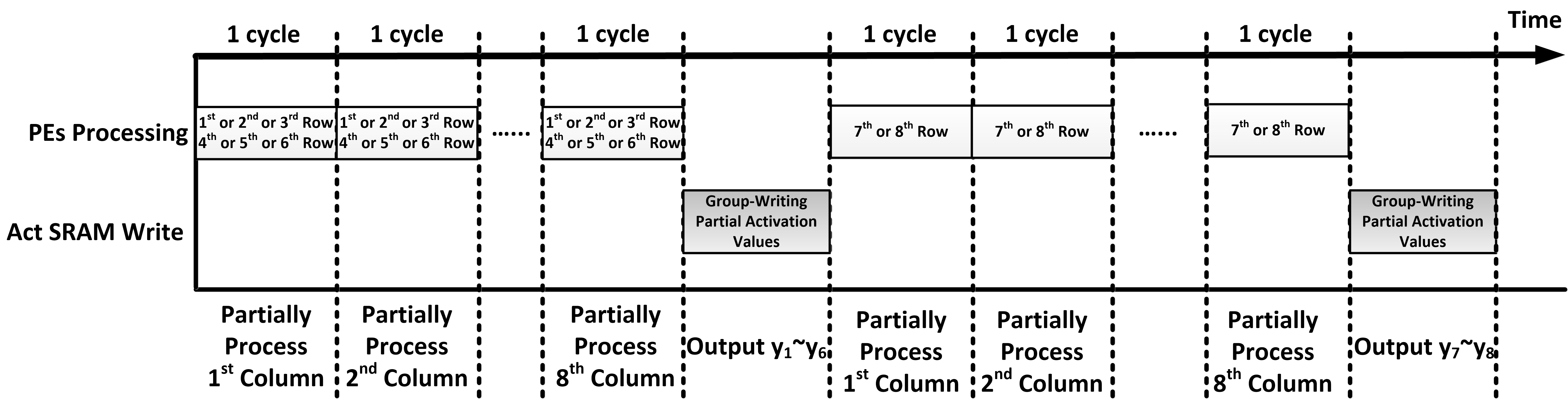}%
\label{fig_second_case}
\caption{Computation scheme with $p=3$.}
\end{subfigure}
\caption{Example computation schemes for a 2-PE \textsc{PermDNN} with $N_{MUL}=1$ and $N_{ACC}=4$ for an 8-by-8 weight matrix.}
\vspace{0mm}
\label{fig:cases}
\end{figure*}

\vspace{0mm}
\textbf{Case 1.} $N_{ROWPE}\geq pN_{MUL}$ and $N_{ACC}\geq N_{ROWPE}$. In this case, since $N_{ACC}\geq N_{ROWPE}$, each PE has sufficient registers to store all the calculated $y_i$'s which it is in charge of. Meanwhile, consider in each cycle one PE processes $N_{MUL}$ size-$p$ permuted diagonal sub-matrices and identifies $N_{MUL}$ non-zero entries among $pN_{MUL}$ rows. Therefore, only $pN_{MUL}$ comparators of one PE are activated to identify the required row indices of non-zero entries. This means that in each accumulation selector of Fig. \ref{fig:acc_sel} only $p$ out of $N_{ACC}/N_{MUL}$ comparators and the corresponding multiplexers are used in each cycle. Therefore, if we denote $N_{ROWPE}=k(pN_{MUL})+d$, where $d\equiv N_{ROWPE}\mod k(pN_{MUL})$, then at the $i$-th cycle of every $k$ cycles, only $(p(i-1)+1)$-th to $(pi)$-th comparators and multiplexers in Fig. \ref{fig:acc_sel} are activated to identify the target register that should be updated. In other words, different $p$ copies of comparators, selectors and accumulators of Fig. \ref{fig:acc_sel} are activated sequentially. Notice that because $N_{ROWPE}=k(pN_{MUL})+d$ and $N_{ACC}\geq N_{ROWPE}$, PEs always have enough hardware resource to continuously process the current column of weight matrix, and it takes $k$ (for $d=0$) or $(k+1)$ (for $d>0$) cycles to fully process one column. For instance, assume we have 2 PEs with $N_{MUL}=1$ and $N_{ACC}=4$ to process an 8-by-8 weight matrix with $p=2$. Then as shown in Fig. \ref{fig:cases}(a), it takes two cycles to process one column and such column-wise processing is continuous.

\vspace{0mm}
\textbf{Case 2.} $N_{ROWPE}\geq pN_{MUL}$ and $N_{ACC}<N_{ROWPE}$. In this case, since $N_{ROWPE}=k(pN_{MUL})+d$, there must exist an integer $f\leq k$ that satisfies $f(pN_{MUL})\leq N_{ACC}<(f+1)(pN_{MUL})$. Therefore, after $f$ cycles of sequential activation described in Case 1, at the $(f+1)$-th cycle the inactivated accumulation selectors and accumulators in Fig. \ref{fig:acc_sel} are not sufficient for future processing. Accordingly, as illustrated in Fig. \ref{fig:cases}(b) with $p=3$, the overall computation scheme needs to be changed for this case: all the other columns need to be first partially processed to calculate part of $y_i$'s. Then the accumulators that were once allocated for the previously calculated $y_i$'s are released and used to perform the rest unfinished computation. Such procedure may need to be repeated for many times if $f$ is much smaller than $k$.


\vspace{0mm}
\textbf{Case 3.} $N_{ROWPE}<pN_{MUL}$. This is a very rare case since the number of PEs should be properly determined to ensure each multiplier is in charge of at least one permuted diagonal sub-matrix. When such case occurs, it means that $p$ is very large (a very sparse model) and some of the PEs become redundant. To solve this problem, we need to change the original single column-processing scheme: different PEs can now process multiple columns simultaneously. Consequently, this strategy leads to two benefits: 1) increase the overall throughput; and 2) equivalently improve $N_{ROWPE}$ for each column processing, and thereby transforming this case to the aforementioned Case 1 or 2.

\vspace{0mm}



\vspace{0mm}
\section{Evaluation}
\label{sec:evaluation}
\subsection{Experimental Methodology}
\label{experiment methods}
\begin{figure}[!h]
\centering
\includegraphics[width=0.9\columnwidth]{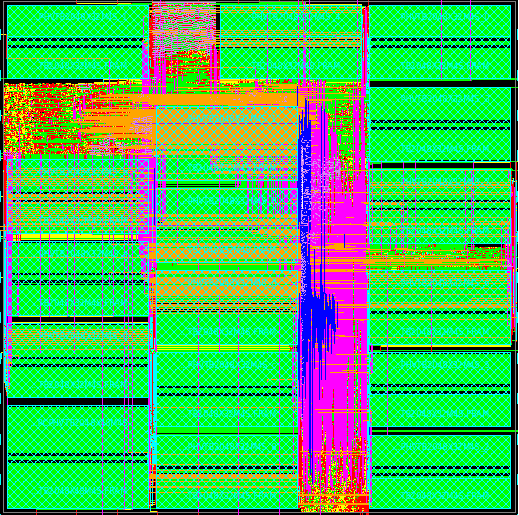}
\caption{Layout of one PE using CMOS 28nm technology.}
\vspace{-3mm}
\label{fig:layout}
\end{figure}

\textbf{Simulation and CAD Tools.}  We developed a cycle-accurate bit-accurate simulator to model the functional behavior of \textsc{PermDNN} architecture. This simulator also serves as the golden reference for the correctness of Verilog implementation. After verifying the RTL model via validating its outputs against the outputs of the simulator, we synthesized our design using Synopsis Design Compiler with CMOS 28nm library. Here the switching activity was extracted from simulation and we annotated the toggle rate to the gate-level netlist. For place and route, we used Synopsis IC compiler to generate layout (see Fig. \ref{fig:layout}). The power consumption was estimated using Prime-Time PX. Notice that the area and power consumption of SRAM part, including weight SRAM, activation SRAM and permutation SRAM, were estimated and reported by Cacti.

\vspace{0mm}\

\textbf{Benchmarks.} Our evaluation uses a set of \textsc{PermDNN} models with the FC layer described in Section \ref{subsec:accuracy}. Specifically, similar to EIE, we evaluate the FC layers of these compressed sparse models individually, and the layers with different sizes are viewed as different workloads. Accordingly, the information of six benchmark layers, including the sizes, constant weight sparsity ratio and statistically calculated activation sparsity ratio\footnote{Lower sparsity ratio means more sparsity.}, are listed in Table \ref{table:workload}.

\begin{table}[h]
\vspace{0mm}
  \caption{Information of evaluated FC layers.}
  \centering
  \begin{tabular}{|c|c|c|c|c|}
    \hline
    \textbf{Layer} & \textbf{Size} &
    \textbf{Weight} & \textbf{Activation} & \textbf{Description} \\ \hline
    Alex-FC6 & \specialcell{4096, 9216} & \specialcell{10\% ($p$=10)} & 35.8\% & \multirowcell{3}{CNN model \\ for image \\ classification} \\\cline{1-4}
    Alex-FC7 & \specialcell{4096, 4096} & \specialcell{10\% ($p$=10)} &  20.6\% & \\\cline{1-4}
    Alex-FC8 & \specialcell{1000, 4096} & \specialcell{25\% ($p$=4)} & 44.4\% &  \\\hline
    NMT-1 & \specialcell{2048,1024} & \specialcell{12.5\% ($p$=8)} & 100\% & \multirowcell{3}{RNN model \\for language \\ translation} \\\cline{1-4}
    NMT-2 & \specialcell{2048, 1536} & \specialcell{12.5\% ($p$=8)} & 100\% & 
    \\\cline{1-4}
    NMT-3\footnote{The weight matrices of NMT model have three types of shapes.} & \specialcell{2048,  2048} & \specialcell{12.5\% ($p$=8)} & 100\% &\\
    \hline
  \end{tabular}
  \vspace{-3mm}
  \label{table:workload}
\end{table}

\subsection{Design Configuration \& Hardware Performance}
\label{subsec:hw performance}
\textbf{Design Configuration.} 
Table \ref{table:design config} lists the design configuration parameters for PE and the overall \textsc{PermDNN} computing engine. Here one PE is equipped with eight 16-bit multipliers and 128 24-bit accumulators. In addition, each PE is also designed to contain $16\times32$ bit$\times2048=128$KB weight SRAM and 48bit$\times2048=12$KB permutation SRAM. It should be noted that the sizes of SRAM are selected by using the “over-design” strategy to ensure the architecture can support a wide range of applications. For instance, by using 4-bit weight sharing strategy a 32-PE \textsc{PermDNN} computing engine can store a compressed layer with 8M parameters, which has the double size of the famous large-size compressed VGG FC6 layer \cite{simonyan2014very}. Similarly, the activation SRAM is also designed with large capacity to facilitate the execution on very large models. As illustrated in Table \ref{table:design config}, the activation SRAM in this design has $8\times64$bit$\times2048=128$KB, which corresponds to a 16-bit 64K-length vector. Such large size is usually sufficient for the FC layers in most practical models.

\begin{table}[t]
\vspace{-0mm}
  \caption{Design configuration parameters.}
  \centering
  \begin{tabular}{|c|c|c|}
    \hline 
    \multicolumn{2}{|c|}{\textbf{PE Parameter}} &  \textbf{Value} \\\hline
    \multirowcell{2}{Multiplier} & Amount ($N_{MUL}$) & 8 \\\cline{2-3}
    & Width & 16 bits \\\hline
    \multirowcell{2}{Accumulator} & Amount ($N_{ACC}$) & 128 \\\cline{2-3}
    & Width & 24 bits \\\hline
    \multirowcell{3}{Weigth SRAM \\ sub-Banks} & Amount & 16 \\\cline{2-3}
    & Width & 32 bits \\\cline{2-3}
    & Depth & 2048 \\\hline
    \multirowcell{2}{Permutation SRAM} & Width & 48 bits \\\cline{2-3}
    & Depth & 2048 \\\hline
    \multicolumn{2}{|c|}{\textbf{\textsc{PermDNN} Computing 
Engine Parameter}} & \textbf{Value} \\\hline
    \multicolumn{2}{|c|}{Amount of PEs ($N_{PE}$)} & 32 \\\hline
    \multicolumn{2}{|c|}{Quantization scheme} & 16 bits \\\hline
    \multicolumn{2}{|c|}{Weight sharing strategy} & 4  bits \\\hline
    \multicolumn{2}{|c|}{Number of pipeline stages} & 5 \\\hline
    \multirowcell{3}{Activation \\SRAM Bank} & Amount ($N_{ACTMB}$) & 8 \\\cline{2-3}
    & Width ($W_{ACTM}$) & 64 bits \\\cline{2-3}
    & Depth & 2048 \\\hline
    \multirowcell{2}{Activation FIFO} & Width & 32 bits \\\cline{2-3}
    & Depth & 32 \\\hline
  \end{tabular}
  \label{table:design config}
  \vspace{0mm}
\end{table}

\textbf{Hardware Performance.} 
Table \ref{table:hw perfomrance} shows the power and area breakdowns for one PE and overall \textsc{PermDNN} computing engine. It is seen that each PE occupies 0.271mm$^2$ and consumes 21.874mW. For the overall 32-PE computing engine, it occupies 8.85mm$^2$ and consumes 703.4mW, where the PE array is the most resource-consuming part (99.5\% power and 97.9\% area).

Regarding the throughput, being equipped with 32 PEs running at 1.2GHz, the \textsc{PermDNN} computing engine can achieve 614.4GOPS for a compressed DNN model. For the equivalent processing power on dense model, since it varies with different compression ratios, we adopt a pessimistic conversion scheme of assuming  8$\times$ weight sparsity and 3$\times$ activation sparsity. Consequently, the equivalent application throughput of \textsc{PermDNN} achieves 14.74TOPS. Notice that  our selected conversion scheme is quite conservative. For instance, EIE adopts conversion scheme of assuming 10$\times$ weight sparsity and 3$\times$ input sparsity, which is more optimistic than ours.
\vspace{0mm}

\begin{table}[h]
  \caption{Power and area breakdowns.}
  \centering
  \begin{tabular}{|c|c|c|}
    \hline 
    \textbf{Component} & \textbf{Power (mW)} & \textbf{Area (mm$^2$)} \\\hline
    \multicolumn{3}{|c|}{\textbf{PE Breakdown}}\\\hline
    Memory & 3.575 (16.35\%) & 0.178 (65.68\%) \\\hline
\specialcell{Register} & 4.755 (21.74\%) & 0.01 (3.69\%) \\\hline
\specialcell{Combinational} & 10.48 (47.91\%) & 0.015 (5.53\%) \\\hline
Clock Network & 3.064 (14.00\%) & 0.0005 (0.18\%)\\\hline \specialcell{Filler Cell} & & 0.0678 (25.02\%) \\\hline
Total & 21.874 & 0.271 \\\hline
& \textbf{Power (mW)} & \textbf{Area (mm$^2$)} \\\hline
    \multicolumn{3}{|c|}{\textbf{\textsc{PermDNN} Computing Engine Breakdown}}\\\hline
32 PEs & 700 & 8.67 \\\hline
Others & 3.4 & 0.18 \\\hline
Total & 703.4 & 8.85 \\\hline
  \end{tabular}
  \vspace{0mm}
  \label{table:hw perfomrance}
\end{table}

\subsection{Comparison with \textsc{EIE} and \textsc{CirCNN}}
\label{subsec:comp with EIE}
In this subsection we compare \textsc{PermDNN} with two most relevant compressed DNN-oriented architectures: EIE and \textsc{CirCNN}. The reasons for selecting these two works as reference are: 1) similar to \textsc{PermDNN}, EIE is a type of FC layer-targeted DNN architecture and it is the state-of-the-art design in this category; 2) similar to \textsc{PermDNN}, \textsc{CirCNN} also imposes structure to the weight matrix of DNN to gain improved hardware performance. 

\textbf{Comparison with EIE.} Table \ref{table:comp eie} summarizes the hardware parameters and performance of EIE and \textsc{PermDNN}. For fair comparison, our design adopts the weight sharing strategy and quantization scheme that are the same to EIE's configuration. Also, because EIE uses a different technology node from ours (45nm vs 28nm), we adopt the scaling strategy used in EIE to project EIE to the same 28nm technology.

\vspace{-0mm}

\begin{table}[h]
\vspace{0mm}
  \caption{Comparison of EIE and \textsc{PermDNN}.}
  \centering
  \begin{tabular}{|c|c|c|c|}
    \hline 
    \textbf{Design} & \multicolumn{2}{|c|}{EIE} & \textbf{\textsc{PermDNN}} \\\hline
    \textbf{Number of PEs} & \multicolumn{2}{|c|}{64} & 32 \\\hline
    \specialcell{\textbf{CMOS Tech.}} & \specialcell{45nm \\(reported)} & 
    \specialcell{28nm \\(projected)\footnote{The projection follows rule in \cite{han2016eie}: linear scaling for frequency, quadratic scaling for area and constant power scaling.} }& 28 nm \\\hline
    \specialcell{\textbf{Clk. Freq. (MHZ)}} & 800 & 1285 & 1200 \\\hline
    \textbf{Memory} & \multicolumn{2}{|c|}{SRAM} & SRAM \\\hline
    \specialcell{\textbf{Weight Sharing}} & \multicolumn{2}{|c|}{4 bits} & 4 bits\footnote{Our experiments show 4-bit weight sharing does not cause accuracy drop.} \\\hline
    \textbf{Quantization} & \multicolumn{2}{|c|}{16 bits} & 16 bits \\\hline
    \textbf{Area (mm$^2$)} & 40.8 & 15.7 & 8.85 \\\hline
    \textbf{Power (W)} & 0.59 & 0.59 & 0.70 \\\hline
  \end{tabular}
  \vspace{0mm}
  \label{table:comp eie}
\end{table}

\begin{figure*}[!h]
\centering
\begin{subfigure}{.3\textwidth}
{\includegraphics[width=\columnwidth]{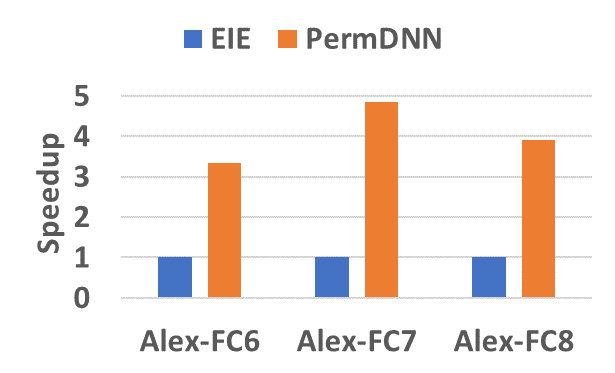}%
\label{fig_comp_speedup}
\vspace{-2mm}
\vspace{-2mm}
}
\end{subfigure}
\begin{subfigure}{.3\textwidth}
{\includegraphics[width=\columnwidth]{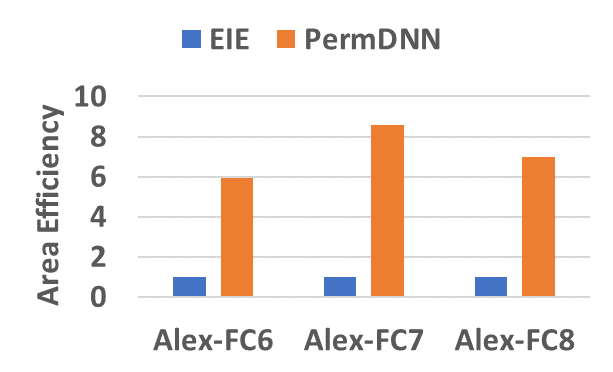}%
\label{fig_comp_area}
\vspace{-2mm}
\vspace{-2mm}
}
\end{subfigure}
\begin{subfigure}{.3\textwidth}
\includegraphics[width=\columnwidth]{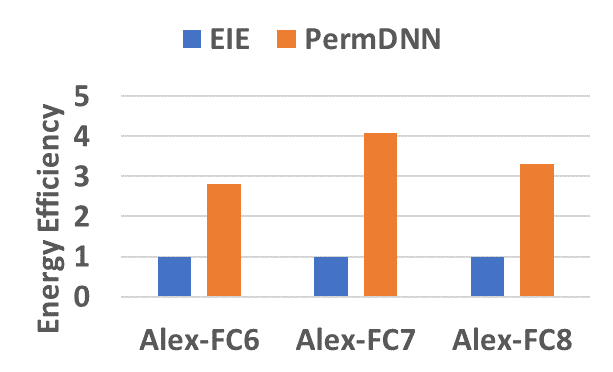}%
\label{fig_comp_energy}
\vspace{-2mm}
\vspace{-2mm}
\end{subfigure}
\caption{Comparison between hardware performance of EIE and \textsc{PermDNN} on different benchmark FC layers.}
\vspace{0mm}
\label{fig:comp}
\end{figure*}

Fig. \ref{fig:comp} compares the hardware performance of EIE and \textsc{PermDNN} on executing three FC layers of AlexNet\footnote{They are the benchmark layers that both EIE and \textsc{PermDNN} evaluate.} in terms of speedup, area efficiency and energy efficiency. It can be seen that compared with the projected EIE using 28nm, \textsc{PermDNN} achieves $3.3\times\sim4.8\times$ higher throughout, $5.9\times\sim8.5\times$ better area efficiency and $2.8\times\sim4.0\times$ better energy efficiency on different benchmark layers. As analyzed in Section \ref{subsec:benefits of pd}, such improvement is mainly due to the elimination of unnecessary storage for weight index, inconvenient address calculation and load imbalance.

\textbf{Comparison with \textsc{CirCNN}.} Table \ref{table:comp circnn} summarizes the hardware performance of \textsc{CirCNN} and \textsc{PermDNN}. Again, for fair comparison we project \textsc{CirCNN} to the same 28nm technology. Notice that because \textsc{CirCNN} reported synthesis results, here in Table \ref{table:comp circnn} the listed performance characteristics of \textsc{PermDNN} are also from synthesis reports.  Meanwhile, because \textsc{CirCNN} only provides power and throughput information, we compare our \textsc{PermDNN} and \textsc{CirCNN} on overall throughput (TOPS) and energy efficiency (TOPS/W). Table \ref{table:comp circnn} shows that \textsc{PermDNN} achieves 11.51$\times$ higher throughput and 3.89$\times$ better energy efficiency. 


As discussed in Section \ref{subsec:permdnn vs cirdnn}, compared with \textsc{CirCNN} the \textsc{PermDNN} enjoys the benefits of utilizing input sparsity and real-number arithmetic computation. Because the architecture of \textsc{PermDNN} and \textsc{CirCNN} are quite different, and the two implementations have their individual design configurations (e.g., compression ratios, size of SRAM, number of multipliers etc.), the contributions of these two advantages to the overall performance improvement are analyzed roughly as follows: 1) Because the utilization of dynamic input sparsity can bring linear reduction in computational cost, if all the other factors that affect throughput and energy efficiency, including compression ratio, clock frequency, number of multipliers and power, were the same for \textsc{PermDNN} and \textsc{CirCNN} designs, $3\times$ input sparsity of \textsc{PermDNN} would enable about $3\times$ increase of throughput and energy efficiency; 2) For the simplicity of real number arithmetic computation, recall that a complex multiplication consists of 4 real multiplications and 2 real additions. Also, notice that performing inference on the same-size compressed weight matrix (e.g., $p$-by-$p$), the permuted diagonal matrix-based approach needs $p$ real multiplications, while circulant matrix-based method requires $p$ complex multiplications in element-wise multiplication phase plus $plogp$ constant complex multiplications in FFT/IFFT phases, Therefore, when the compression ratio is the same, the use of real-number arithmetic computation can roughly bring about $4\times$ reduction in computational cost, which can translate to a significant improvement in hardware performance.

\vspace{0mm}
\subsection{Scalability}
\label{subsec:scale flex}
\begin{figure}[!h]
\centering
\includegraphics[width=1\columnwidth]{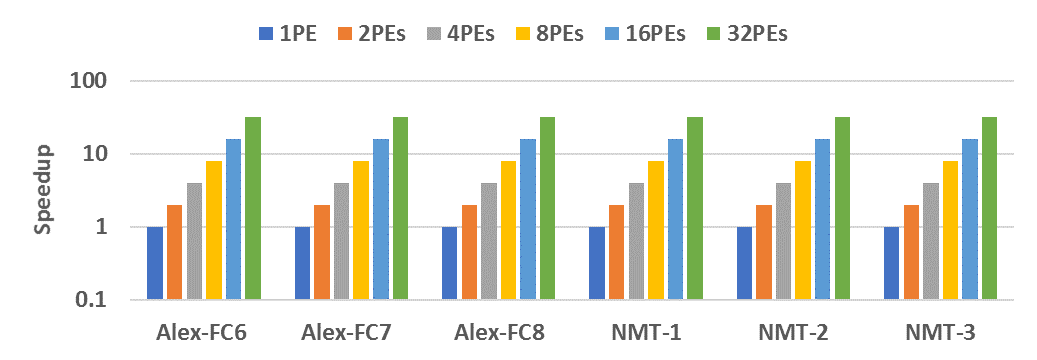}
\caption{Scalability of \textsc{PermDNN} on different benchmarks.}
\vspace{0mm}
\label{fig:scalability}
\end{figure}

\vspace{0mm}
As the size of weight matrix of FC layer grows, it is easy to scale up \textsc{PermDNN} by adding more PEs. In general, thanks to block-permuted diagonal matrix's unique characteristic of even distribution of non-zero entries along row and column directions, \textsc{PermDNN} enjoys strong scalability since the load imbalance problem \cite{han2016eie} that challenges other sparse DNN accelerators does not exist for \textsc{PermDNN} at all. Fig. \ref{fig:scalability} shows the speedup of \textsc{PermDNN} architecture using different numbers of PEs. We see that our design achieves very good scalability on all benchmarks. 

\vspace{0mm}

\begin{table}[h]
\vspace{0mm}
  \caption{Comparison of \textsc{CirCNN} and \textsc{PermDNN}. Both results are from synthesis reports.}
  \centering
  \begin{tabular}{|c|c|c|c|}
    \hline 
    \textbf{Design} & \multicolumn{2}{|c|}{\textsc{CirCNN}} & \textbf{\textsc{PermDNN}} \\\hline
    \textbf{Number of PEs} & \multicolumn{2}{|c|}{N/A} & 32 \\ \hline
    \specialcell{\textbf{CMOS Tech.}} & \specialcell{45 nm \\(reported)} & 
    \specialcell{28 nm \\(projected)} & 28 nm \\ \hline
    \specialcell{\textbf{Clk. Freq. (MHZ)}} & 200 & 320 & 1200 \\\hline
    \textbf{Quantization} & \multicolumn{2}{|c|}{16 bits} & 16 bits \\\hline
    \textbf{Area (mm$^2$)} & N/A & N/A & 6.64 \\\hline
    \textbf{Power (W)} & 0.08 & 0.08 & 0.236 \\\hline
    \specialcell{\textbf{Throughput} \\ \textbf{(Equivalent TOPS)}} & 0.8 & 1.28 & \specialcell{\textbf{14.74}\\ \textbf{(11.51$\times$)} }\\\hline
    \specialcell{\textbf{Energy Efficiency} \\ \textbf{(Equivalent TOPS/W)}} & 10.0 & 16.0 & \specialcell{\textbf{62.28}\\\textbf{(3.89$\times$)}}\\\hline
  \end{tabular}
  \vspace{0mm}
  \label{table:comp circnn}
\end{table}

\section{Related Work}
\label{sec:related work}
Besides EIE and \textsc{CirCNN}, various works have been reported for exploring high-performance DNN design. Diannao family \cite{chen2014diannao,du2015shidiannao,liu2015pudiannao,chen2014dadiannao,liu2016cambricon,zhang2016cambricon} and Google's TPU \cite{jouppi2017datacenter} propose a series of DNN processors and the corresponding instruction sets. In \cite{albericio2016cnvlutin,chen2016eyeriss,parashar2017scnn,yu2017scalpel,hill2017deftnn,reagen2016minerva,leestitch}, several sparse DNN accelerators are investigated. Considering the importance of meomory in high-performance DNN hardware, \cite{kim2016neurocube,chi2016prime,li2017drisa,gao2017tetris,rhu2016vdnn} extensively study better utilization of memory in DNN.

Among the research effort on data flow in DNN, \cite{lu2017flexflow,maeri_asplos2018,alwani2016fused} propose optimized data flow to provide high flexibility or reduce memory access. Also, \cite{judd2016stripes,albericio2017bit,ren2017sc} investigate high-performance DNN hardware using bit-serial inputs. Besides, various FPGA-based DNN designs have been implemented and reported, including \cite{sharma2016high,shen2017maximizing,qiu2016going,shen2016overcoming,zhao2017accelerating,moss2018customizable,moss2017high}.

Beyond the research works on digital accelerators for inference tasks, DNN hardware using analog or mixed-signal circuits \cite{shafiee2016isaac,likamwa2016redeye,bojnordi2016memristive,song2017pipelayer,bankman2018always,feinberg2018making} are also the potential high-efficiency solutions for deploying deep learning. Considering training is a very time-consuming process, several architecture-level approaches are proposed in \cite{venkataramani2017scaledeep,de2017understanding,rhu2018compressing,song2018towards,song2018situ,mahajan2016tabla} to improve the efficiency of training.

DNN model compression is also an active research topic in machine learning community. Various approaches, including pruning \cite{han2015learning}, low rank decomposition \cite{jaderberg2014speeding}, quantization \cite{lin2016fixed}\cite{choi2018universal}, structured matrix \cite{cheng2015exploration} and clustering \cite{gong2014compressing}, have been proposed in prior works. \cite{cheng2017survey} presents a survey of different compression techniques.

\vspace{-2mm}
\section{Conclusion}
\label{sec:conclusion}
This paper proposes \textsc{PermDNN}, a novel approach to generate and execute hardware-friendly structured sparse DNN models using permuted diagonal matrices. Compared with prior model compression approaches, \textsc{PermDNN} enjoys the benefits of regularity, flexibility and simple arithmetic computation. We then propose \textsc{PermDNN} architecture, a scalable FC layer-targeted computing engine. Implementation results show \textsc{PermDNN} achieves $3.3\times\sim4.8\times$ higher throughout, $5.9\times\sim8.5\times$ better area efficiency and $2.8\times\sim4.0\times$ better energy efficiency than EIE. It also achieves $11.51\times$ higher throughput and $3.89\times$ better energy efficiency than \textsc{CirCNN}.

\section{Acknowledgement}
\label{sec:acknowledgement}
The authors would like to appreciate anonymous reviewers' valuable comments and suggestions. This work is funded by the National Science Foundation Awards CCF-1815699, CCF-1814759, CNS-1717984 and CCF-1750656.


\bibliographystyle{IEEEtran.bst}
\bibliography{ref}

\end{document}